\documentclass{article}

\usepackage[final,nonatbib]{neurips_2024}

\usepackage[utf8]{inputenc} %
\usepackage[T1]{fontenc}    %
\usepackage{hyperref}       %
\usepackage{url}            %
\usepackage{booktabs}       %
\usepackage{amsfonts}       %
\usepackage{amsmath}
\usepackage{mathtools}
\usepackage{bm}
\usepackage{nicefrac}       %
\usepackage{microtype}      %
\usepackage{xcolor}         %
\usepackage{graphicx}
\usepackage[noabbrev,capitalise]{cleveref}
\usepackage{comment}
\usepackage[group-separator={,}]{siunitx}

\usepackage[frozencache,cachedir=./_minted-paper]{minted}
\usepackage{multirow}

\usepackage[
    style=authoryear,
    maxcitenames=2,
    mincitenames=1,
    maxbibnames=99,
    uniquelist=minyear,
    uniquename=init,
    giveninits,
    dashed=false,
]{biblatex}
\DeclareNameAlias{sortname}{family-given}
\AtEveryBibitem{\clearfield{month}}
\AtEveryBibitem{\clearfield{day}}
\AtEveryBibitem{\ifentrytype{unpublished}{}{\clearfield{eprint}\clearfield{eprinttype}}}
\DeclareSourcemap{
    \maps[datatype=bibtex, overwrite]{
        \map{
            \pertype{article}
            \pertype{inproceedings}
            \pertype{unpublished}
            \step[fieldset=url, null]
            \step[fieldset=urldate, null]
        }
    }
}
\DeclareSourcemap{
	\maps[datatype=bibtex, overwrite]{
		\map{
			\step[fieldset=doi, null]
			\step[fieldset=issn, null]
			\step[fieldset=isbn, null]
			\step[fieldset=editor, null]
			\step[fieldset=publisher, null]
			\step[fieldset=location, null]
			\step[fieldset=organization, null]
			\step[fieldset=series, null]
		}
	}
}
\newcommand{\method}[1]{\textsc{#1}}
\newcommand{\Tk}{Top-$k$}
\newcommand{\tk}{top-$k$}
\newcommand{\E}{\mathbb{E}}
\def\lc{\left\lceil}
\def\rc{\right\rceil}
\def\lf{\left\lfloor}
\def\rf{\right\rfloor}

\usepackage{xcolor}

\usepackage{tikz}
\usepackage[most]{tcolorbox}
\usetikzlibrary{arrows.meta}
\usetikzlibrary{calc}
\usetikzlibrary{patterns,arrows,decorations.pathreplacing}
\usetikzlibrary{shapes.geometric}
\definecolor{flare1of3}{rgb}{0.92907237, 0.68878959, 0.50411509}
\definecolor{flare2of3}{rgb}{0.75861834, 0.25356035, 0.40663694}
\definecolor{flare3of3}{rgb}{0.29408557, 0.13721193, 0.38442775}
\definecolor{flare1of5}{HTML}{8E336F}
\definecolor{flare2of5}{HTML}{AF3D6B}
\definecolor{flare3of5}{HTML}{CF4B62}
\definecolor{flare4of5}{HTML}{E16960}
\definecolor{flare5of5}{HTML}{E88D70}

\definecolor{gc_pink}{HTML}{FF6F79}
\definecolor{gc_blue}{HTML}{6FB0FF}
\definecolor{gc_gray}{HTML}{D9D9D9}
\definecolor{gc_dark_pink}{HTML}{99262E}
\definecolor{gc_dark_blue}{HTML}{265A99}
\definecolor{gc_dark_gray}{HTML}{999999}
\definecolor{comment_color}{HTML}{1B8F44}
\definecolor{comment_color_2}{RGB}{64,128,128}

\newtcbox{\scenariobox}[1][]{enhanced,
 box align=base,
 nobeforeafter,
 colback=gc_gray,
 size=small,
 left=1pt,
 right=1pt,
 top=-0.2pt,
 bottom=0.3pt,
 boxsep=1.2pt,
 boxrule=0pt,
 #1}
\newcommand{\casesmallk}{\scenariobox{$k \ll n$}}
\newcommand{\casepropk}{\scenariobox{$k \propto n$}}

\title{Approximate Top-$\bm{k}$ for Increased Parallelism}

\newcommand{\authorspacing}{\,\,\,\,\,\,}

\author{%
    Oscar Key$^\textbf{1}$ \authorspacing Luka Ribar$^\textbf{2}$ \authorspacing Alberto Cattaneo$^\textbf{2}$ \authorspacing Luke Hudlass-Galley$^\textbf{2}$ \authorspacing Douglas Orr$^\textbf{2}$\\
      $^\text{1}$Centre for Artificial Intelligence, University College London \,\,\,\,\,\, $^\text{2}$Graphcore Research\\
      \texttt{oscar.key.20@ucl.ac.uk} \,\,\,\,\, \texttt{\{lukar, albertoc, lukehg, douglaso\}@graphcore.ai} \\
}

\begin{document}

\maketitle

\begin{abstract}
    We present an evaluation of bucketed approximate \tk{} algorithms.
    Computing \tk{} exactly suffers from limited parallelism, because the $k$ largest values must be aggregated along the vector,  thus is not well suited to computation on highly-parallel machine learning accelerators.
    By relaxing the requirement that the \tk{} is exact, bucketed algorithms can dramatically increase the parallelism available by independently computing many smaller \tk{} operations.
    We explore the design choices of this class of algorithms using both theoretical analysis and empirical evaluation on downstream tasks.
    Our motivating examples are sparsity algorithms for language models, which often use \tk{} to select the most important parameters or activations.
    We also release a fast bucketed \tk{} implementation for PyTorch.

\end{abstract}
\begin{figure}[h]
    \centering
    \includegraphics[width=\textwidth]{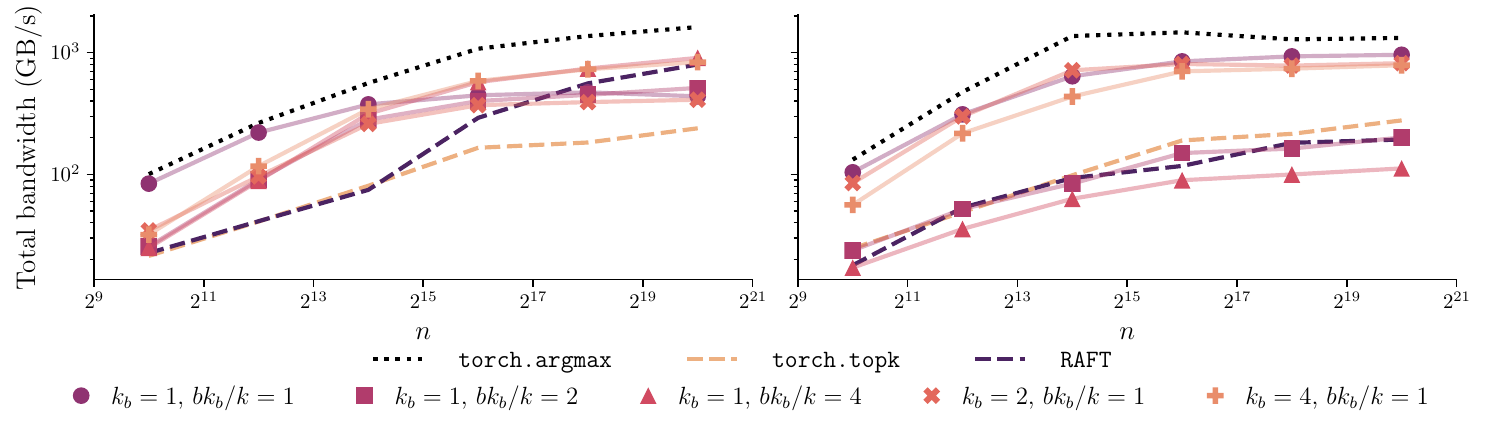}
    \vspace*{-1.5em}
    \caption[Speed of our implementation compared to exact top-ks]{
    Our approximate \tk{} implementation
    (\tikz[baseline=-0.5ex]{
        \node[circle, fill=flare1of5, inner sep=2.0pt] at (0,0) {};
        \node[rectangle, fill=flare2of5, inner sep=2.5pt] at (0.3,0) {};
        \node[regular polygon, regular polygon sides=3, fill=flare3of5, inner sep=1.5pt] at (0.6,-0.02) {};
        \node[text=flare4of5, scale=1.2, font=\bfseries] at (0.9,0) {×};
        \node[text=flare5of5, scale=1.2, font=\bfseries] at (1.2,0) {+};
    }\hspace{-3pt}),
    compared with exact \tk{} implementations from PyTorch and \texttt{RAFT},
    and a bucketed \tk{} using \texttt{torch.argmax}, tested in \texttt{float32} on an H100 PCIe GPU with batch size $m=128$.
    Total bandwidth is the minimum number of bytes transferred by \tk{}, divided by runtime.
    \emph{Left:} Small fixed $k=64$; it is faster to retrieve $k_b=1$ element per bucket, varying the total number $b\cdot k_b$ of elements retrieved, where $b$ is the number of buckets. \emph{Right:} Large $k=n/4$; best to set $b\cdot k_b/k=1$ and increase $k_b$.
    }
    \label{fig:headline-performance}
    \vspace*{-1em}
\end{figure}

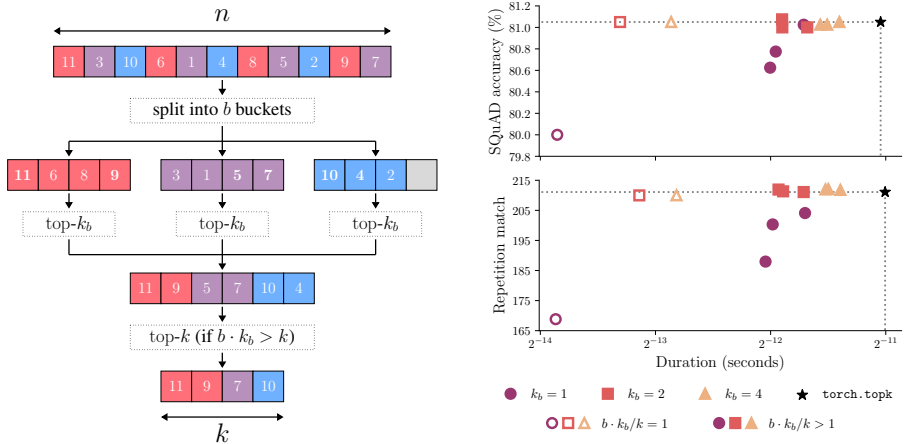
\begin{figure*}[t]
    \centering
    \begin{minipage}[t]{0.45\textwidth}
        \resizebox{\columnwidth}{!}
        {

\definecolor{p01}{HTML}{EB9B74}
\definecolor{p02}{HTML}{E98768}
\definecolor{p03}{HTML}{E5715E}
\definecolor{p04}{HTML}{DE5D5C}
\definecolor{p05}{HTML}{D34C60}
\definecolor{p06}{HTML}{C14168}
\definecolor{p07}{HTML}{AE3B6D}
\definecolor{p08}{HTML}{9A3670}
\definecolor{p09}{HTML}{863071}
\definecolor{p10}{HTML}{722C6E}
\definecolor{p11}{HTML}{5E2868}

\definecolor{b1}{HTML}{FF6F79}
\definecolor{b2}{HTML}{B790BC}
\definecolor{b3}{HTML}{6FB0FF}
\definecolor{gray}{HTML}{D9D9D9}

\begin{tikzpicture}

\fill[fill=none] (0, 14) rectangle (11, 14) node[pos=0.5] {\Huge $n$};
\draw[very thick, Triangle-Triangle, color=black] (0, 13.5) -- (11, 13.5);

\filldraw[draw=black, fill=b1] (0, 12) rectangle (1, 13) node[pos=0.5] {\Large\textcolor{white}{$11$}};
\filldraw[draw=black, fill=b2] (1, 12) rectangle (2, 13) node[pos=0.5] {\Large\textcolor{white}{$3$}};
\filldraw[draw=black, fill=b3] (2, 12) rectangle (3, 13) node[pos=0.5] {\Large\textcolor{white}{$10$}};
\filldraw[draw=black, fill=b1] (3, 12) rectangle (4, 13) node[pos=0.5] {\Large\textcolor{white}{$6$}};
\filldraw[draw=black, fill=b2] (4, 12) rectangle (5, 13) node[pos=0.5] {\Large\textcolor{white}{$1$}};
\filldraw[draw=black, fill=b3] (5, 12) rectangle (6, 13) node[pos=0.5] {\Large\textcolor{white}{$4$}};
\filldraw[draw=black, fill=b1] (6, 12) rectangle (7, 13) node[pos=0.5] {\Large\textcolor{white}{$8$}};
\filldraw[draw=black, fill=b2] (7, 12) rectangle (8, 13) node[pos=0.5] {\Large\textcolor{white}{$5$}};
\filldraw[draw=black, fill=b3] (8, 12) rectangle (9, 13) node[pos=0.5] {\Large\textcolor{white}{$2$}};
\filldraw[draw=black, fill=b1] (9, 12) rectangle (10, 13) node[pos=0.5] {\Large\textcolor{white}{$9$}};
\filldraw[draw=black, fill=b2] (10, 12) rectangle (11, 13) node[pos=0.5] {\Large\textcolor{white}{$7$}};

\draw[very thick, -Triangle, color=black] (5.5, 11.9) -- (5.5, 11.4);

\filldraw[draw=black,dotted,fill=none] (2.5, 11.3) rectangle (8.5, 10.5) node[pos=0.5] {\LARGE split into $b$ buckets};

\draw[very thick, -, color=black] (5.5, 10.4) -- (5.5, 9.9);

\draw[very thick, -, color=black] (0.5, 9.9) -- (10.5, 9.9);
\draw[very thick, -Triangle, color=black] (0.5, 9.9) -- (0.5, 9.4);
\draw[very thick, -Triangle, color=black] (5.5, 9.9) -- (5.5, 9.4);
\draw[very thick, -Triangle, color=black] (10.5, 9.9) -- (10.5, 9.4);

\filldraw[draw=black, fill=b1] (-1.5, 8.3) rectangle (-0.5, 9.3) node[pos=0.5] {\Large\textcolor{white}{$\bm{11}$}};
\filldraw[draw=black, fill=b1] (-0.5, 8.3) rectangle (0.5, 9.3) node[pos=0.5] {\Large\textcolor{white}{$6$}};
\filldraw[draw=black, fill=b1] (0.5, 8.3) rectangle (1.5, 9.3) node[pos=0.5] {\Large\textcolor{white}{$8$}};
\filldraw[draw=black, fill=b1] (1.5, 8.3) rectangle (2.5, 9.3) node[pos=0.5] {\Large\textcolor{white}{$\bm{9}$}};

\filldraw[draw=black, fill=b2] (3.5, 8.3) rectangle (4.5, 9.3) node[pos=0.5] {\Large\textcolor{white}{$3$}};
\filldraw[draw=black, fill=b2] (4.5, 8.3) rectangle (5.5, 9.3) node[pos=0.5] {\Large\textcolor{white}{$1$}};
\filldraw[draw=black, fill=b2] (5.5, 8.3) rectangle (6.5, 9.3) node[pos=0.5] {\Large\textcolor{white}{$\bm{5}$}};
\filldraw[draw=black, fill=b2] (6.5, 8.3) rectangle (7.5, 9.3) node[pos=0.5] {\Large\textcolor{white}{$\bm{7}$}};

\filldraw[draw=black, fill=b3] (8.5, 8.3) rectangle (9.5, 9.3) node[pos=0.5] {\Large\textcolor{white}{$\bm{10}$}};
\filldraw[draw=black, fill=b3] (9.5, 8.3) rectangle (10.5, 9.3) node[pos=0.5] {\Large\textcolor{white}{$\bm{4}$}};
\filldraw[draw=black, fill=b3] (10.5, 8.3) rectangle (11.5, 9.3) node[pos=0.5] {\Large\textcolor{white}{$2$}};
\filldraw[draw=black, fill=gray] (11.5, 8.3) rectangle (12.5, 9.3);

\draw[very thick, -Triangle, color=black] (0.5, 8.2) -- (0.5, 7.7);
\draw[very thick, -Triangle, color=black] (5.5, 8.2) -- (5.5, 7.7);
\draw[very thick, -Triangle, color=black] (10.5, 8.2) -- (10.5, 7.7);

\filldraw[draw=black,dotted,fill=none] (-1, 7.6) rectangle (2, 6.8) node[pos=0.5] {\LARGE $\mathrm{top}\text{-}k_b$};
\filldraw[draw=black,dotted,fill=none] (4, 7.6) rectangle (7, 6.8) node[pos=0.5] {\LARGE $\mathrm{top}\text{-}k_b$};
\filldraw[draw=black,dotted,fill=none] (9, 7.6) rectangle (12, 6.8) node[pos=0.5] {\LARGE $\mathrm{top}\text{-}k_b$};

\draw[very thick, -, color=black] (0.5, 6.7) -- (0.5, 6.2);
\draw[very thick, -, color=black] (5.5, 6.7) -- (5.5, 6.2);
\draw[very thick, -, color=black] (10.5, 6.7) -- (10.5, 6.2);
\draw[very thick, -, color=black] (0.5, 6.2) -- (10.5, 6.2);
\draw[very thick, -Triangle, color=black] (5.5, 6.2) -- (5.5, 5.7);

\filldraw[draw=black, fill=b1] (2.5, 5.6) rectangle (3.5, 4.6) node[pos=0.5] {\Large\textcolor{white}{$11$}};
\filldraw[draw=black, fill=b1] (3.5, 5.6) rectangle (4.5, 4.6) node[pos=0.5] {\Large\textcolor{white}{$9$}};
\filldraw[draw=black, fill=b2] (4.5, 5.6) rectangle (5.5, 4.6) node[pos=0.5] {\Large\textcolor{white}{$5$}};
\filldraw[draw=black, fill=b2] (5.5, 5.6) rectangle (6.5, 4.6) node[pos=0.5] {\Large\textcolor{white}{$7$}};
\filldraw[draw=black, fill=b3] (6.5, 5.6) rectangle (7.5, 4.6) node[pos=0.5] {\Large\textcolor{white}{$10$}};
\filldraw[draw=black, fill=b3] (7.5, 5.6) rectangle (8.5, 4.6) node[pos=0.5] {\Large\textcolor{white}{$4$}};

\draw[very thick, -Triangle, color=black] (5.5, 4.5) -- (5.5, 4);
\filldraw[draw=black,dotted,fill=none] (2.5, 3.9) rectangle (8.5, 3.1) node[pos=0.5] {\LARGE $\mathrm{top}\text{-}k$ (if $b \cdot k_b > k$)};

\draw[very thick, -Triangle, color=black] (5.5, 3) -- (5.5, 2.5);

\filldraw[draw=black, fill=b1] (3.5, 2.4) rectangle (4.5, 1.4) node[pos=0.5] {\Large\textcolor{white}{$11$}};
\filldraw[draw=black, fill=b1] (4.5, 2.4) rectangle (5.5, 1.4) node[pos=0.5] {\Large\textcolor{white}{$9$}};
\filldraw[draw=black, fill=b2] (5.5, 2.4) rectangle (6.5, 1.4) node[pos=0.5] {\Large\textcolor{white}{$7$}};
\filldraw[draw=black, fill=b3] (6.5, 2.4) rectangle (7.5, 1.4) node[pos=0.5] {\Large\textcolor{white}{$10$}};

\fill[fill=none] (0, 0.4) rectangle (11, 0.4) node[pos=0.5] {\Huge $k$};
\draw[very thick, Triangle-Triangle, color=black] (3.5, 0.9) -- (7.5, 0.9);

\end{tikzpicture}}
    \end{minipage}
    \hspace{1em}
    \begin{minipage}[t]{0.41\textwidth}
        \resizebox{\columnwidth}{!}
        {\includegraphics{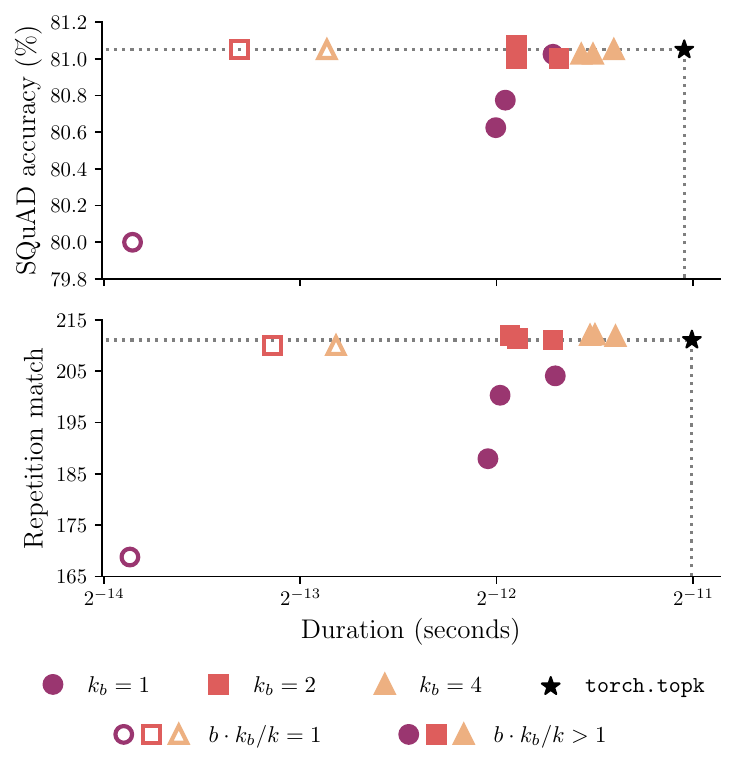}}
    \end{minipage}
    \vspace*{-0.3em}
    \caption{
        \emph{Left:} An example of a bucketed \tk{}, with $n=11$, $k=4$, $b=3$ and $k_b=2$. In \textbf{Stage 1}, $n$ elements are reduced to $b \cdot k_b$ elements via $b$ independent top-$k_b$. An optional \textbf{Stage 2} takes final top-$k$.
        \emph{Right:} The trade-off between \tk{} runtime duration and downstream task accuracy for SparQ Attention in SQuAD and a sequence repetition task (see \cref{app:downstream-task-details}), when using different bucketed \tk{} settings with batch size $m=1$. Good accuracy and speed-ups above $4\times$ are achieved with $k_b\in\{2,4\}$.
    }
    \label{fig:schematic}
    \label{fig:trade-off-sparq}
    \vspace*{-1em}
\end{figure*}

\section{Motivation and existing work}
\Tk{} is an operation, common to most machine learning frameworks, that selects the $k$ largest elements from a vector of length $n$.
It appears in many machine learning algorithms, for example $k$-nearest neighbours (KNN) similarity search, and top-$k$ sampling in language models \parencite{fan2018topksampling}.
Of particular interest are its application to sparsity methods, often aimed at reducing the otherwise substantial costs of large language models (LLMs).
These methods ignore part of the parameters or activations in order to reduce the amount of computation, either during training, during generation \parencite{sheng2023flexgen,ribar2024sparq}, or both \parencite{jayakumar2020topkast}.
Many sparsity methods discard the parameters or activations with the smallest magnitude, and implement this selection using \tk{}.
Thus, these algorithms may call \tk{} frequently and on long vectors.
For example, \textcite{jayakumar2020topkast} applies \tk{} to each weight matrix of a Transformer during all training iterations, with each matrix potentially containing millions of parameters.
Similarly, \textcite{sheng2023flexgen,ribar2024sparq} compute \tk{} over the sequence length, which might comprise tens of thousands of tokens, in each attention layer and for each token generated.

However, for larger $n$, computing \tk{} can be slow: our experiments with \textcite{ribar2024sparq} (\Cref{sec:e2e-speedups}) show that \tk{} consumes $>10$\% of the time to generate each token and is not able to fully utilise compute resources.
One reason for this is that parallelism over $n$ is necessary --- the batch dimension is rarely large enough to saturate the processor --- but this is limited because substantial cooperation is required between threads.
A common class of methods partitions the input vector into buckets, discards those buckets that fall outside the \tk{}, and recurses into buckets that are partially within the \tk{}.
Parallelism is achieved by using many threads for the partitioning stage, however the buckets must be shared.
Within this class, \textcite{alabi2012gpuselection} introduce \method{BucketSelect} and \method{RadixSelect}, as used by PyTorch \parencite{paszke2019pytorch}; see also \textcite{ribizel2019sampleselect, zhang2023benchmarking}.
Another approach is to use multiple threads to scan over the input while maintaining a shared priority queue of the $k$ largest items \parencite{johnson2021warpselect, zhang2023benchmarking}.
\method{bitonic top-k} \parencite{shanbhag2018efficienttopk} and \method{Dr. Top-k} \parencite{gaihre2021drtopk} are hierarchical approaches, which split the input into buckets, perform selection in each bucket in parallel, and then, possibly recursively, merge the buckets until the final \tk{} is obtained.
Merging the buckets again requires cooperation between threads, limiting parallelism.

More parallelism can be achieved by computing an approximate \tk{}.
This approach is particularly relevant for sparsity algorithms because \tk{} is generally already a heuristic, thus an additional approximation may not impact the overall performance.
In this paper we consider methods which split the input into buckets (a common strategy for parallel reduction, see e.g. \textcite{tritonlayernorm}), perform a smaller top-$k_b$ in each bucket ($k_b \le k$), and combine the outputs.
This design is similar to the hierarchical exact algorithms.
However, as the output must no longer be exact, less or no cooperation is required when combining the output of the buckets.
An existing instance is \method{TPU-KNN} \parencite{chern2022tpu}, implemented for TPUs as the \texttt{approx\_max\_k} operation in JAX and TensorFlow.
Their design choices are tailored to KNN applications where $k \ll n$, for example $n \sim$ millions and $k < 100$.
We find that these choices are not optimal for sparsity applications, where $k$ might be $5\%$ - $20\%$ of $n$.
While the \texttt{FAISS} library contains several GPU-efficient approximate KNN implementations, these mostly rely on an exact \tk{}.
A single approximate \tk{} method is documented \parencite{guzhva2023faissapprox}, which only supports CPU (and for which we haven't found any evaluations).

We aim to make fast approximate \tk{} algorithms more readily available to practitioners by investigating the design choices of bucketed \tk{}, and by releasing our implementation as a PyTorch library\footnote{\url{https://github.com/graphcore-research/pytorch-approx-topk}}.
We evaluate cost-quality trade-offs of the algorithm using theoretical analysis and empirical experiments on a sparse attention method for LLMs, LLM vocabulary sampling, and knowledge graph completion.
We demonstrate that substantial speed-ups of the \tk{} operation are possible in many settings.
In particular, on the sparse attention method task, we observe the \tk{} to be over $4 \times$ faster with little to no degradation in downstream task performance.

\section{Algorithm design choices}
\label{sec:algorithm}
We consider bucketed algorithms consisting of two stages (see \cref{fig:schematic}, left):
\begin{description}
    \item[Stage 1] divide the $n$ inputs into $b$ interleaved buckets, select the largest $k_b$ values in each bucket, and concatenate the $b \cdot k_b$ results;
    \item[Stage 2] if $b \cdot k_b > k$, use an exact \tk{} to select the $k$ largest values.
\end{description}
The design parameters are $b \in \{1, \ldots, n\}$ and $k_b \in \{1, \ldots, \min(k, n/b)\}$, with the requirement that $b \cdot k_b \geq k$.
\method{TPU-KNN} is an instance where $k_b = 1$ and \textbf{Stage 2} is implemented using a bitonic sort.
The implementation in \texttt{FAISS} uses $b = k / k_b$, thus \textbf{Stage 2} is not required.

The quality of the approximation can be improved by increasing either $b$ or $k_b$.
We consider two regimes: \casesmallk{}, $k$ fixed to a small value (common in KNN applications); \casepropk{}, $k$ is a significant proportion of $n$ (common in sparsity applications).
In \casesmallk{}, we might expect that the computation time is dominated by \textbf{Stage 1}, thus our intuition is to improve quality by increasing $b$ so that more parallelism is available.
In \casepropk{}, the cost of \textbf{Stage 2} becomes significant, thus quality should be improved by increasing $k_b$ and maintaining $b \cdot k_b = k$.
Our analysis supports these intuitions, as shown in the following sections.

We could use any exact \tk{} algorithm in \textbf{Stage 1}.
However, in our empirical evaluations we found that good approximation and efficiency can be achieved by capping $k_b \leq 4$.
We can take advantage of this property to write a fast implementation that uses a small priority queue stored in registers and requires no or minimal communication between threads.
See \cref{app:algorithm-details}.

\section{Theoretical evaluation} \label{sec:theoretical_eval}
\begin{figure}[t]
    \centering
    \includegraphics[width=\textwidth]{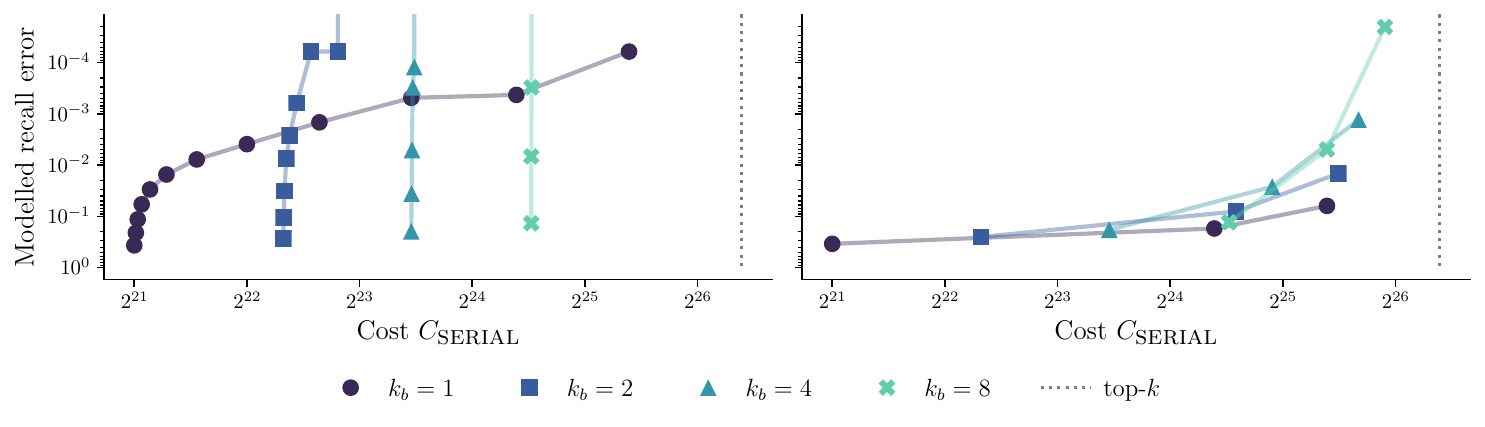}
    \vspace*{-1em}
    \caption{Theoretical trade-off curves, using the serial cost model, which computes the count of all operations executed in an abstract execution model (\cref{app:cost-models}), for $n=\num{1048576}$ ($=2^{20}$), with \casesmallk{} (\emph{left}, $k\!=\!256$) and \casepropk{} (\emph{right}, $k\!=\!n/8\!=\!\num{131072}$). Points along the curves (from bottom to top) indicate increasing the $b \cdot k_b/k$ ratio, leading to a decreasing recall error. See \cref{appendix:tradeoff-curves} for the full set of trade-off curves with various cost models.}
    \label{fig:trade-off-regimes}
\end{figure}

As an initial evaluation of the design choices in an implementation-neutral manner, we use theoretical models to investigate the trade-off between the quality of the approximation and computational cost.
We measure quality using recall, $\bm{R}(k,b,k_b) = \bm{Z}(k,b,k_b)/k$, where the random variable $\bm{Z}(k,b,k_b)$ counts the number of true \tk{} values returned by the approximate method, obtaining an upper bound on the expected recall error, $1 - \E[\bm{R}]$, see \cref{eq:theoretical_recall_uniform}, and derive several theoretical cost models for different implementations based on operation counting, see \cref{app:cost-models} for details.

\Cref{fig:trade-off-regimes} shows the trade-off between recall error and theoretical cost, as we vary $b$ and $k_b$ under the two regimes.
The lowest point on each curve represents the case $b \cdot k_b = k$, and moving along the curve (from bottom to top) corresponds to increasing $b$ further.
Both regimes show that it is theoretically possible to achieve substantial speed-ups without much loss in recall.
For \casesmallk{}, we observe that increasing $b$ alone is the most cost-effective way of improving recall.
For \casepropk{}, larger values of $b \cdot k_b$ incur a substantial performance penalty due to the cost of \textbf{Stage 2}, thus it is better to increase recall by choosing a larger value of $k_b$ and a smaller value of $b$.

\section{Empirical evaluation}
Our empirical evaluation consists of benchmarking speed of our bucketed \tk{} implementation against exact \tk{}, exploring the cost-quality trade-off in downstream tasks, and demonstrating that using an approximate \tk{} can achieve end-to-end speed-ups when generating from an LLM.
Further details are presented in \cref{app:experiment-details}.

\subsection{Runtime benchmarks}
\Cref{fig:headline-performance} compares the memory bandwidth achieved by our implementation, under various configurations, with three baselines:
\texttt{torch.argmax} (an approximate \tk{} implemented by bucketing the PyTorch argmax function), \texttt{torch.topk}, and the highly GPU-efficient exact method from the \texttt{RAFT} library \parencite{rapidsai}, which implements \method{Air Top-K} and \method{GridSelect} as developed by \textcite{zhang2023benchmarking}.

\texttt{torch.argmax} implements bucketed \tk{} for the case $b = k$, $k_b=1$.
If we assume that it is well optimised, it constitutes an upper bound for bucketed methods, showing opportunities for substantial performance gains of exact \tk{}, particularly in the case \casepropk{}.
Our implementation for the case $b = k$, $k_b=1$ exhibits good scaling, but would require further optimisation to reach the performance of the upper bound, especially for \casesmallk{}.
However, for most choices of parameters, our implementation offers substantial speed-ups over the exact methods, except for \casesmallk{} when $n$ is large where \texttt{RAFT} is particularly fast.

Note that \texttt{RAFT} only supports \texttt{float32} values, thus this is the data type we use in \cref{fig:headline-performance}.
Language modelling tasks commonly use lower precision data types, so we also compare the bandwidth for \texttt{bfloat16} in \cref{fig:bfloat16-bandwidth}, though the ranking of the methods is largely the same.
The subsequent evaluations on downstream tasks are performed with \texttt{bfloat16}, thus we only compare against \texttt{torch.topk}.

\subsection{Cost-quality trade-off}
\begin{figure}[t]
    \centering
    \includegraphics[width=\textwidth]{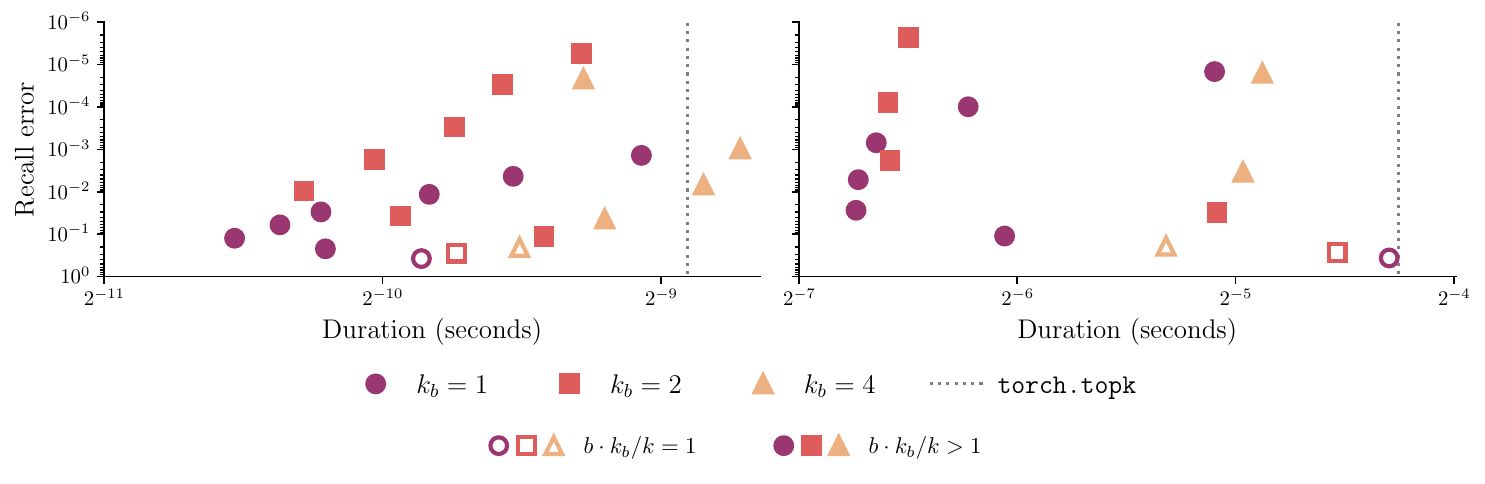}
    \vspace*{-1em}
    \caption{
    Bucketed \tk{} trade-off for LLM vocabulary sampling (\emph{left}, $n=\num{128256}$, $k=256$, $m=64$) and for knowledge graph link prediction (\emph{right}, $n=\num{2653751}$, $k=100$, $m=128$).
    In both regimes, $k_b=1$ gives peak performance, but $k_b=2$ sacrifices some speed for sake of a lower error and is Pareto optimal when increasing $b$.
    }
    \label{fig:trade-off-downstream}
\end{figure}

\paragraph{Sparse attention in LLM inference}
We consider SparQ Attention \parencite{ribar2024sparq}, which speeds up generation for long sequences in Transformer models.
\Tk{} is applied to the attention scores to select which values should be fetched from the KV cache.
Here \casepropk{}, with $n/16 \leq k \leq n/8$, where $n$ is the sequence length.
\Cref{fig:trade-off-sparq} shows the cost-quality curves for Llama $3$ $8$B \parencite{dubey2024llama} on two evaluation tasks.
Consistent with our theoretical results, increasing $k_b$ is the most cost-effective way of improving task performance.
In fact, for $k_b = 2$ and $b \cdot k_b / k = 1$, the cost of \tk{} is reduced by more than $4\times$, with almost no degradation in quality.

\paragraph{Small $\bm{k}$ tasks}
We consider two tasks where \casesmallk{}.
A common application of \tk{} in LLMs is to select the subset of most likely next tokens to sample from during generation.
We again perform our evaluation on Llama $3$ $8$B, which has a vocabulary size $n=\num{128256}$, and choose $k=256$.
Finally, we look at link-prediction on knowledge graphs, another very common application of \tk{}.
We investigate tail predictions on PharMeBINet \parencite{konigs2022_pharmebinet}, a large biomedical knowledge graph with $n=2.6 \times 10^6$ entities, and $k=100$.
\Cref{fig:trade-off-downstream} shows the recall error (compared to exact \tk{}) for these two tasks.
Consistent with the theoretical results, we observe that increasing $k_b$ and $b \cdot k_b / k$ together can retain good recall while achieving speed-ups between $2\times$ and $4\times$.

\subsection{End-to-end speed-ups} \label{sec:e2e-speedups}
\begin{figure}
    \centering
    \includegraphics[width=\textwidth]{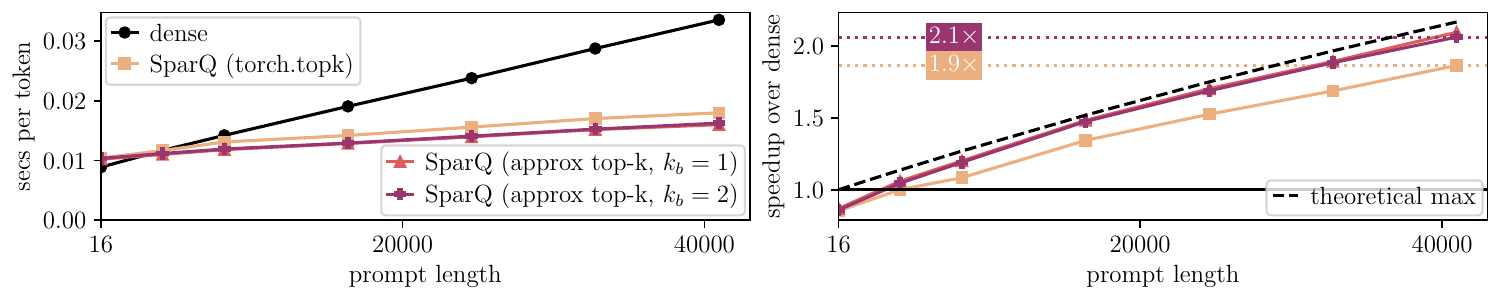}
    \caption[End-to-end speed-ups for LLM text generation]{
        End-to-end speed-ups achieved when generating text from Llama 2 7B using SparQ sparse attention, which relies on the \tk{} operation.
        We use $k=n/8$ and batch size $1$, and set $b$ such that $b \cdot k_b = k$.
        \tikz[baseline=-0.5ex]{\draw [line width=0.5mm, dashed] (0,0) -- (2ex,0);} shows the theoretical maximum speed-up that SparQ could achieve if the computation was fully memory-bandwidth-bound (see \cref{app:e2e-speedup-details}).
        We plot the mean over four repeats, observing little variance.
    }
    \label{fig:sparq-e2e-speedup}
\end{figure}
We return to SparQ Attention, as introduced above, to demonstrate that using a bucketed \tk{} can achieve end-to-end speed-ups when generating from an LLM.
Here we consider Llama 2 7B and batch size $1$, and measure the time taken to generate the $n+1$th token as we increase the prompt length, $n$.
\Cref{fig:sparq-e2e-speedup} shows the results.
As expected, the speed-up offered by SparQ over dense attention is small for shorter prompts --- in fact it is slower for the shortest prompts due to overheads --- but substantial for longer prompts.
This is because the time spent in the attention operation grows linearly with the length of the prompt while other costs remain constant, thus a faster attention algorithm has a greater effect on the total time.
We observe that replacing the exact \tk{} in SparQ with a bucketed \tk{} results in a modest additional performance boost: from $1.9 \times$ faster than dense to $2.1 \times$ faster (\textasciitilde $ 10\%$) for a prompt of $40,000$ tokens.

\section{Conclusions}
We have shown that additional parallelism can be introduced to the \tk{} operation with a bucketing approach without affecting downstream task performance.
Notably, different strategies are optimal depending on the ratio $k/n$: sparsity applications are better served by top-$k_b$ ($k_b>1$) per bucket, whereas if $k\ll n$ increasing the number of buckets is more cost-effective.
In most cases our implementation provides substantial speed-ups over exact \tk{}, except in the case of \casesmallk when $n$ is large, where RAFT also has very good performance.
Two limitations of our evaluation are that we have not benchmarked end-to-end performance outside \textcite{ribar2024sparq}, and have not thoroughly investigated the effect of batch size.
Future work could also examine the performance of bucketed methods in distributed settings, where the \tk{} is performed by multiple accelerators.
Here, bucketed methods may have a larger advantage due to the limited communication required between buckets.
Finally, we hope that the PyTorch library that we release will make it easier for sparsity researchers to start using approximate \tk{} algorithms in their work.

\begin{ack}
    OK acknowledges support from the Engineering and Physical Sciences Research Council with grant number EP/S021566/1.
\end{ack}

\printbibliography

\clearpage
\appendix

\section{Algorithm and implementation details} \label{app:algorithm-details}
We release a CUDA implementation of bucketed \tk{}.
The implementation has two modes to ensure enough parallelism is exposed: (1) $1$ thread per bucket, (2) $64$ threads per bucket. In the benchmarks presented in the paper, for each specific configuration of $n, k, b, k_b$ we always select the mode, between the two, giving the best performance.
The implementation also offers a heuristic to select the mode: use mode (1) if
\begin{itemize}
    \item number of threads $\ge$ number of lanes on the device,
    \item or, buckets contain $<64$ elements.
\end{itemize}
In mode (1), each thread selects the top-$k_b$ values from each bucket and writes them back to memory.
In mode (2), each thread selects the top-$k_b$ values from a subsection of a bucket, the resulting values are sorted within each group of $64$ threads, and the final top-$k_b$ values are written to memory.
The merge is performed using the BlockMergeSort primitive from the CUB library \parencite{nvidiacccl}.
This is a cheap operation because it can be performed in shared memory and only takes place once.
In both modes the implementation of each thread is the same: the thread iterates over the input while inserting each value into a max priority queue of size $k_b$.
As $k_b \leq 4$, the priority queue can be cheaply and simply implemented using insertion sort in per-thread registers.

Mode (1) is the same as the per-thread \tk{} described by \textcite{shanbhag2018efficienttopk}.
Mode (2) is similar to \method{BlockSelect} \parencite{johnson2021warpselect}, except we can take advantage of $k_b \leq 4$ to only merge the per-thread queues once, as opposed to periodically having to do it during the computation.
Note that both of these previous implementations store the queue in fast but small memories, namely the registers or shared memory, thus they are limited to values of $k < 1024$.
The bucketed approach can support much larger values of $k$ while still storing the queues in fast memories.

\subsection{Bucket assignment} \label{app:interleaving}
When splitting $n$ inputs into $b$ buckets ($b \le n$), we investigate two different ways of assigning elements along the sequence to buckets: \textit{contiguous} and \textit{interleaved}. Let $\mathcal{B}: \left\{ 0, \ldots, n-1 \right\} \rightarrow \left\{0, \ldots, b-1 \right\}$ be the assignment function, so that input $x_i$ is assigned to bucket $\mathcal{B}(i)$.
\begin{itemize}
\item In  \textit{contiguous} mode, we set $\mathcal{B}(i) := \lf \frac{bi}{n} \rf$.
\item In \textit{interleaved} mode, we set $\mathcal{B}(i) := i \pmod{b}$.
\end{itemize}

\begin{figure}[p]
    \centering
    \includegraphics[width=\textwidth]{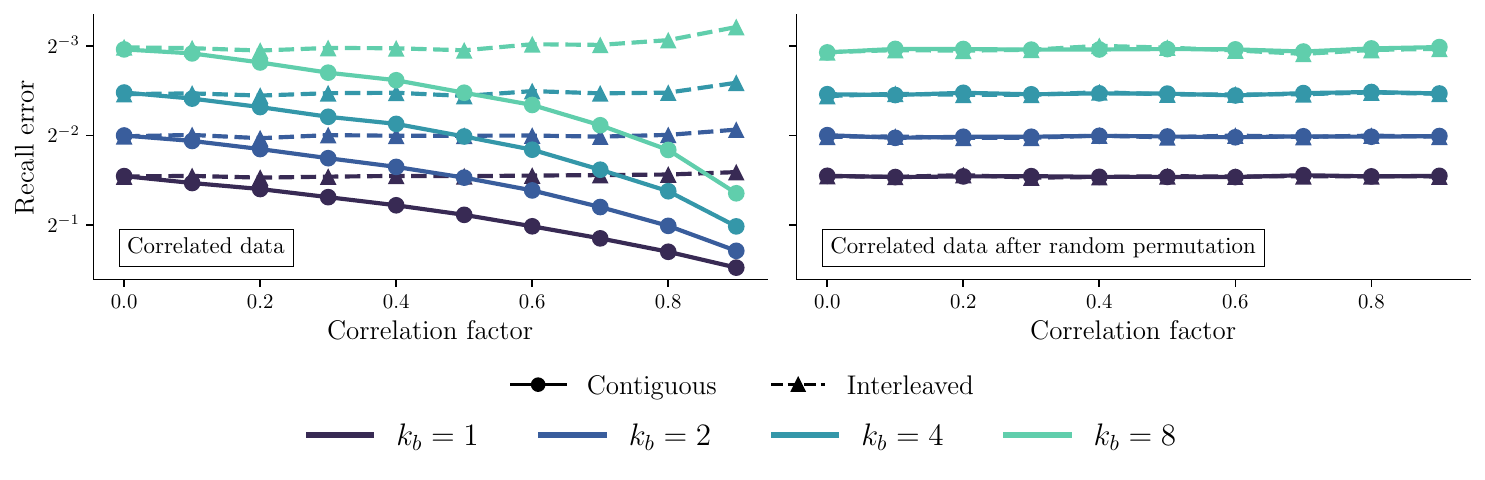}
    \vspace*{-1em}
    \caption{
    Comparison of achieved recall vs. exact \tk{} using interleaved and contiguous bucket assignments on normally distributed correlated data.
    \emph{Left:} As the correlation between data points is increased, contiguous assignment showcases significant recall degradation, even as $k_b$ is increased.
    \emph{Right:} When the correlated data is randomly permuted, the contiguous assignment achieves the same recall as the interleaved assignment.}
    \label{fig:appendx-correlation-sim}
\end{figure}

An ideal bucket assignment would break any correlations in the input data, so that each bucket can be expected to contain approximately the same number of \tk{} values as any other (uniform distribution assumption; see \cref{app:recall-model}). However, data is often highly correlated across the sequence, which can lead to poor recall when using contiguous assignment compared to exact \tk{}. We simulate this effect by repeatedly drawing a sequence of $n = 2048$ elements from a Multivariate normal distribution $\boldsymbol{x} \sim \mathcal{N} (0, \boldsymbol{\Sigma})$ such that $\Sigma_{i, j} = \rho^{|i - j|}$ with the correlation factor $\rho \in [0, 1)$. Applying the approximate bucket \tk{} algorithm with $k = 256$ for various $k_b$, we observe that, as the correlation between the elements is increased, the contiguous assignment leads to dramatic degradation in recall; on the other hand, interleaved assignment works very well in retaining recall, no matter the degree of correlation (\cref{fig:appendx-correlation-sim}, left). In cases when permuting the data can be done cheaply, the performance can be alternatively retained by first shuffling the data (\cref{fig:appendx-correlation-sim}, right).

\begin{figure}[p]
    \centering
    \includegraphics[width=\textwidth]{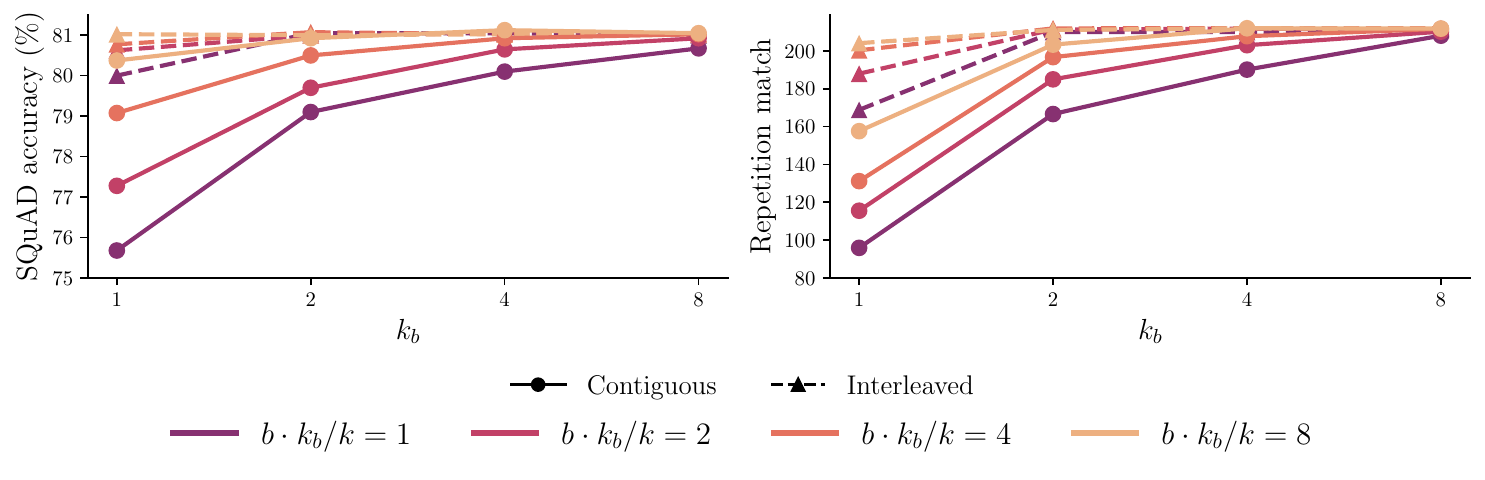}
    \vspace*{-1em}
    \caption{
    Comparison of LLM downstream task performance using interleaved and contiguous bucket assignment with SparQ attention. As textual sequences are highly correlated, there is a significant performance degradation when using contiguous assignment.
    \emph{Left:} SQuAD task.
    \emph{Right:} Repetition task.
    }
    \label{fig:appendix-correlation-sparq}
\end{figure}

\begin{figure}[p]
    \centering
    \includegraphics[width=\textwidth]{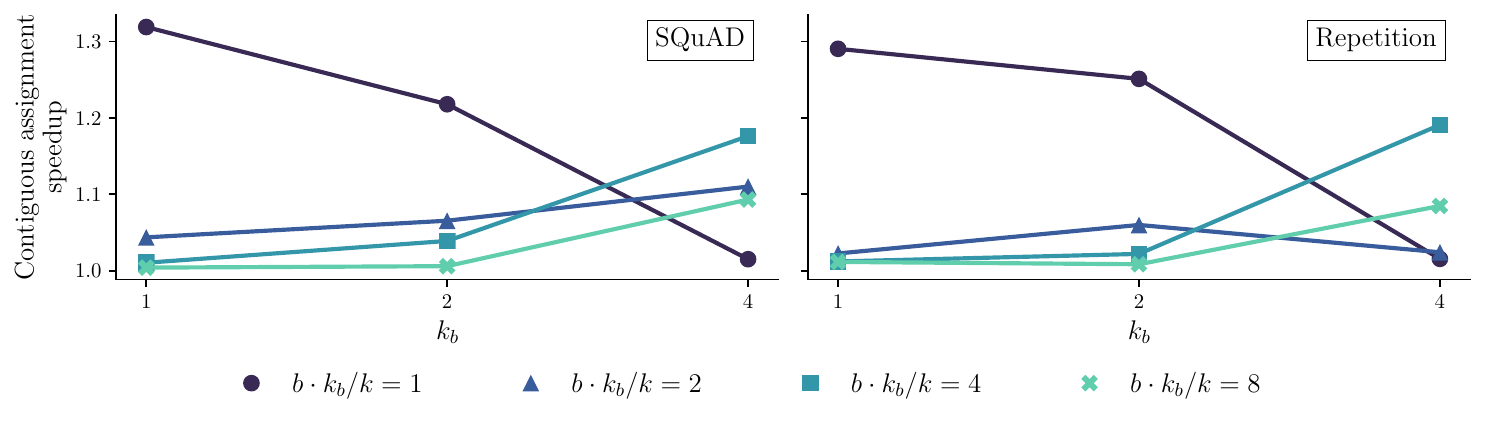}
    \vspace*{-1em}
    \caption{Runtime comparison of contiguous vs interleaved assignment on SQuAD and Repetition tasks, indicating the speed-up of the contiguous assignment.}
    \label{fig:appendix-correlation-speedup-sparq}
\end{figure}

All the real-world data that we tested in this work contains significant correlation between the ordered values -- for example, \cref{fig:appendix-correlation-sparq} showcases the performance degradation when using contiguous assignment for the LLM downstream task with SparQ attention. On the other hand, due to the contiguity of memory accesses, the contiguous assignment showcases runtime improvements compared to the interleaved assignment, as can be seen in \cref{fig:appendix-correlation-speedup-sparq} using the same \tk{} settings as in the LLM downstream task experiments. Nevertheless, due to the data correlation issues, we assign the elements to buckets in the \textit{interleaved} fashion throughout our experiments.

\section{Experiment details}
\label{app:experiment-details}
\subsection{Runtime benchmarks}
Benchmark results were all generated on a H100 PCIe GPU, using PyTorch 3.12 and CUDA 12.1.
Our benchmark performs the following steps:
\begin{enumerate}
    \item Define tensor $x$ of shape $(m, n)$
    \item Warmup loop ($16$ iterations):
    \begin{enumerate}
        \item Fill $x$ with i.i.d. values from a unit normal distribution
        \item Launch \tk{} kernel
    \end{enumerate}
    \item Timing loop ($512$ iterations):
    \begin{enumerate}
        \item Fill $x$ with i.i.d. values from a unit normal distribution
        \item Launch sleep kernel
        \item Record start CUDA event
        \item Launch \tk{} kernel
        \item Record stop CUDA event
    \end{enumerate}
\end{enumerate}
By launching the sleep kernel within the timing loop we ensure the start/stop events and the \tk{} kernel have all been queued before the first start event is executed, thus avoiding including Python overhead in our timing measurements.
We find that this setup yields very stable timings.
Over all the benchmarking runs presented in this work, maximum ratio of standard error to mean duration is $5\%$, and the average ratio is $0.1\%$.

\begin{figure}[t]
    \centering
    \includegraphics[width=\textwidth]{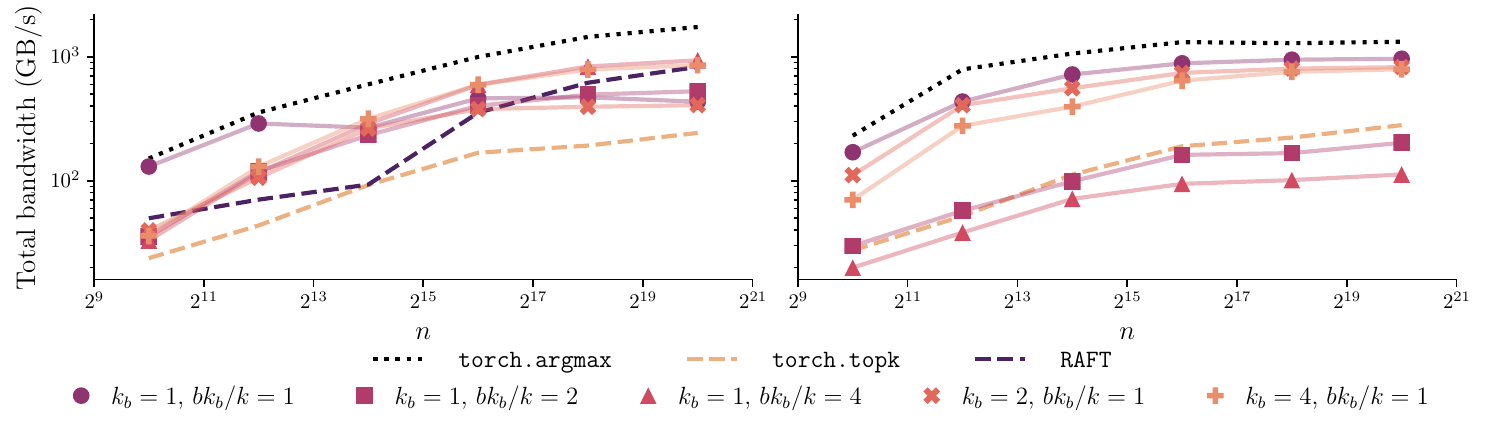}
    \vspace*{-1.5em}
    \caption{
        Bandwidth comparison with CUDA graphs enabled, in the same configurations used for \cref{fig:headline-performance}.
        The data type is \texttt{float32}.
        \emph{Left:} Small fixed $k=64$. \emph{Right:} Large $k=n/4$.
    }
    \label{fig:cuda-graphs-bandwidth-bandwidth}
    \bigbreak
    \includegraphics[width=\textwidth]{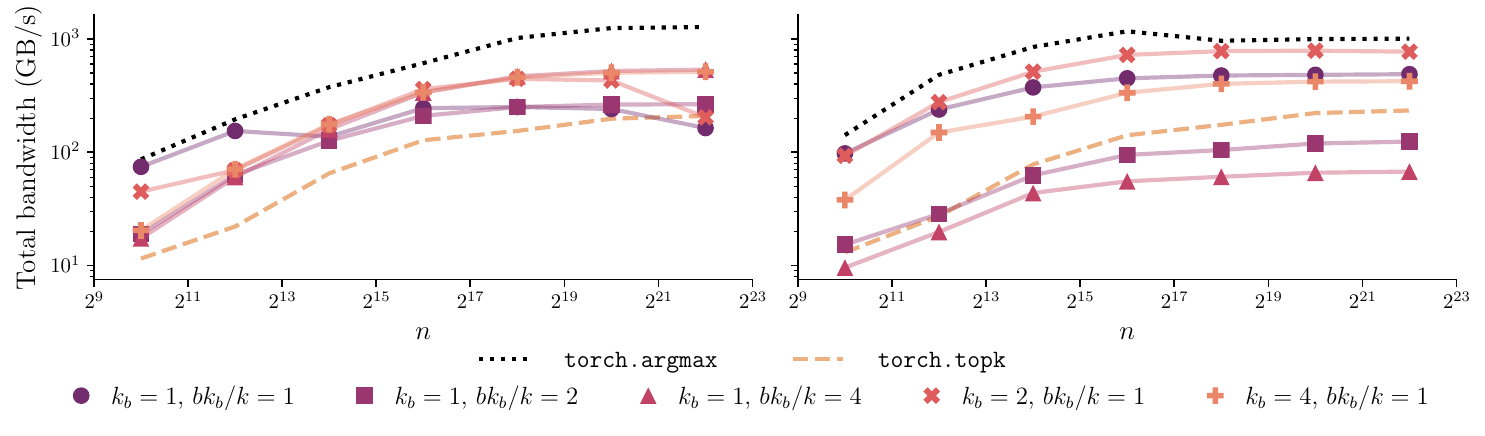}
    \vspace*{-1.5em}
    \caption{
    Bandwidth comparison for \texttt{bfloat16} values, in the same configurations used for \Cref{fig:headline-performance}.
    RAFT is not included because it only supports \texttt{float32}.
    CUDA graphs are enabled.
    \emph{Left:} Small fixed $k=64$. \emph{Right:} Large $k=n/4$.
    }
    \label{fig:bfloat16-bandwidth}
\end{figure}

Ideally we would use CUDA graphs \parencite{gray2019_cudagraphs} to improve the accuracy of the timings by reducing overheads due to memory allocations and CPU/GPU communication, however these are not compatible with the radix \tk{} implementation in RAFT (used when $k>256$).
To investigate the impact of this on the results, \cref{fig:cuda-graphs-bandwidth-bandwidth} shows the working configurations with CUDA graphs enabled.
We do not observe any difference in the ordering of the methods, thus decide to present the results without CUDA graphs so RAFT can be fully included.

\subsection{Cost-quality trade-off} \label{app:downstream-task-details}
\subsubsection{LLM generation with SparQ Attention}
SparQ Attention \parencite{ribar2024sparq} is an attention-approximating algorithm for speeding up long-sequence transformer generation. It consists of two main steps. Firstly, an approximate dot-product of the current query and the cached keys is conducted, generating approximate scores across the sequence length. Secondly, \tk{} approximate scores are selected, and attention is approximated using only these $k$ key-value pairs.

We examine the effect of using approximate \tk{} for selecting top key-value pairs within \textbf{Step 2} of the SparQ Attention algorithm. The \tk{} algorithm is applied to the approximate attention scores, and $k$ is generally chosen to be $1/16$ - $1/8$ of the sequence length. We test the effectiveness of the approximation by evaluating its effect on the downstream task performance on two representative tasks taken from \parencite{ribar2024sparq}: question answering on SQuAD dataset and text repetition. For the former, we use a modified version of SQuAD v1.1 \parencite{rajpurkar2016squad} where each provided context (required to answer the question) is augmented with seven additional unrelated contexts sampled from other examples.
For the latter, the model is prompted to repeat a piece of text from its context, and the task performance is measured by the length of the exact match of the answer in characters.
We perform both evaluations using the Llama $3$ $8$B model \parencite{dubey2024llama}.

\subsubsection{Knowledge graph completion}
Knowledge Graphs (KGs) encode domain knowledge in the form of subject-relation-object triples $(h,r,t)$, with $h, t$ in the set $\mathcal{V}$ of graph nodes (also known as \textit{entities}) and $r$ the edge type (or \textit{relation type}). As KGs are often incomplete, inferring missing links is a classical ML application, which in turn can be framed in terms of triple completion, e.g. finding the most likely tail for a query $(h,r,?)$. Knowledge Graph Embedding models learn embeddings for entities and relation types, which are used as inputs to a real-valued scoring function $f(\mathbf{h}, \mathbf{r}, \mathbf{t})$ whose output is interpreted as a likelihood score for the triple $(h,r,t)$. Finding the set of $k$ most likely completions for a query $(h,r,?)$ becomes then a \tk{} problem: $\text{$k$-argmax}_{t \in \mathcal{V}} f(\mathbf{h}, \mathbf{r}, \mathbf{t})$. We refer to \textcite{kge_survey} for more details.

In this task $n = \left| \mathcal{V} \right|$ can be very large (in the order of tens of millions for real-world KGs), while $k$ is typically fixed and small ($k \le 100$ in most practical applications and model benchmarks, see \textcite{hu2020_ogb}). For our experiments, we use PharMeBINet \parencite{konigs2022_pharmebinet}, a biomedical KG with $\num{2653751}$ entities, and consider the task of predicting tails on a random held-out test set of $\num{55000}$ triples. As KGE model, we take the basic -- but widely used \parencite{hu2020_ogb} -- DistMult scoring function $f(\mathbf{h}, \mathbf{r}, \mathbf{t}) = \langle \mathbf{r}, \mathbf{h}, \mathbf{t} \rangle$ \parencite{yang2015_embedding}. Training and inference are performed using the BESS distribution framework \parencite{cattaneo2022_bess}.

\subsection{End-to-end speed-ups} \label{app:e2e-speedup-details}
As above, we apply the approximate \tk{} to Step 2 of the SparQ algorithm.
We select a compression ratio of $1/8$, hence $k=n/8$.

To perform the experiments we modify the ``gpt-fast'' implementation of SparQ (\url{https://github.com/graphcore-research/sparq-gpt-fast}), releasing the modifications here:
\url{https://github.com/oscarkey/sparq-gpt-fast-approx-topk} (archive DOI 10.5281/zenodo.14205917).
We use an Nvidia H100 PCIe, using PyTorch 2.4 with CUDA 12.4 and cuDNN 9.

\Cref{fig:sparq-e2e-speedup} shows the theoretical maximum speed-up that SparQ could achieve, under the assumption that LLM inference is purely memory bound.
This calculation is available in the file \texttt{theoretical\_speedups.py}, function \texttt{speedup\_theoretical\_time\_in\_attn()}.
We note that SparQ with approximate \tk{} achieves closer to this maximum speed-up, as it is more compute intensive than exact \tk{} and thus more memory bound.

\section{Recall model}
\label{app:recall-model}

As in \cref{sec:theoretical_eval}, let $\bm{Z}(k,b,k_b)$ be the number of true \tk{} values returned by the approximate method, and  $\bm{R}(k,b,k_b) := \bm{Z}(k,b,k_b)/k$ be the recall. For the sake of modelling, we first assume that the input data is uniformly distributed, $k \ge k_b$ and $k_b \ll n/b$. Then:
\begin{equation*}
    \E[\bm{Z}(k,b,k_b)] = \E[\bm{Z}(k-1,b,k_b)]\phi_k + (\E[\bm{Z}(k-1,b,k_b)]+1)(1-\phi_k) = \E[\bm{Z}(k-1,b,k_b)] + 1 - \phi_k,
\end{equation*}
where we denote by $\phi_k$ the probability that the $k$-th top value falls in a bucket already containing at least $k_b$ of the top-($k-1$) values.
As top values are equally likely to be contained in any of the buckets, $\phi_k = 1 - F(k_b-1;k-1,1/b)$, where $F(x;m,p) := \mathrm{Pr}(X \le x)$ is the cumulative function of the binomial distribution $X \sim \mathrm{Binom}(m,p)$. Moreover, $\E[\bm{Z}(k_b,b,k_b)] = k_b$, hence by recursion the \textbf{expected recall error} is:
\begin{equation}\label{eq:theoretical_recall_uniform}
    1 - \E[\bm{R}(k,b,k_b)] = 1 - \frac{1}{k}\left(k_b + \sum_{i=k_b}^{k-1}F\left(k_b-1;i,\frac{1}{b}\right)\right).
\end{equation}

We point out that, when $k_b=1$:
\begin{equation*}
    \E[\bm{R}(k,b,1)] = \frac{1}{k}\left(1 + \sum_{i=1}^{k-1}\left( \frac{b-1}{b}\right)^i\right) = \frac{b}{k}\left( 1 - \left(\frac{b-1}{b} \right)^k\right)
\end{equation*}
differs from the expected recall computed in the TPU-KNN paper \parencite{chern2022tpu}: indeed, the authors of ibid.\ assume that only the values (among the \tk{}) that do not share a bucket with any other ones contribute to $\bm{Z}(k,b,1)$, which is an overly-pessimistic hypothesis (if two \tk{} values fall in the same bucket, one of them is still retrieved by the approximate method; equivalently, $\bm{Z}(k,b,1)$ coincides with the number of unique buckets containing a \tk{} value).

When dropping the assumption $k_b \ll n/b$, we observe that if a bucket contains an excess of top-($k$-1) values compared to its peers, it will be less likely to contain the $k$-th top value too (differently from the previous analysis where all buckets were always considered equiprobable, based on the assumption that only a negligible number of values needed to be retrieved from each of them). As a consequence, in this more general setting \cref{eq:theoretical_recall_uniform} can still be seen as a upper bound for the expected recall error.

If we also remove the hypothesis of uniformly distributed inputs, the worst-case scenario happens when all true \tk{} values are concentrated in the smallest possible number of buckets, that is $\lc \frac{bk}{n} \rc$. For this worst-case, the recall error is:
\begin{equation} \label{eq:theoretical_recall_worst}
1 - \bm{R}_{\text{wc}}(n,k,b,k_b) = 1 - \frac{1}{k}\left( \lf \frac{bk}{n}\rf k_b + \min\left(k_b, k \; (\mathrm{mod}\ \frac{n}{b}) \right)\right).
\end{equation}
From experiments on real-world data, however, we find that the recall error achieved by our implementation (using interleaved bucket assignment to break correlations in the sequence, see \Cref{app:interleaving}) is always sufficiently closely aligned with \cref{eq:theoretical_recall_uniform}, while \cref{eq:theoretical_recall_worst} gives too pessimistic bounds; therefore, we do not include the latter in our discussion.

\section{Cost models}
\label{app:cost-models}

To evaluate the efficiency of the approximate \tk{} algorithm in a general setting, independent of a specific hardware platform and implementation, we devise a set of abstract \emph{cost models} for exact and approximate \tk{} algorithms. These serve as a guide — simplifying the complexities of real hardware and software platforms, they provide an understanding of the algorithmic factors influencing performance and the different regimes that we can expect to see in practical benchmarks. After describing the models in this section, we consider their agreement against GPU benchmarks in \ref{app:cost-models-analysis}.

Each cost model contains multiple algorithms, and the overall cost $C_{\mathcal{M}}(n, k, m)$ from a model $\mathcal{M}$ is the minimum across the algorithms $\mathcal{A}$ supported by that model, $C_{\mathcal{M}} = \min_{\mathcal{A}} C_{\mathcal{M},\mathcal{A}}$. Cost models for exact \tk{} are summarised in \cref{tab:cost-models}.

Given a cost model for exact \tk{}, we can form a model for the approximate \tk{} algorithm described in \cref{sec:algorithm} as
\begin{equation*}
    \tilde{C}_{\mathcal{M}}(n, k, m, b, k_b) =
    C_{\mathcal{M}}(n/b, k_b, m \cdot b)
    + [b \cdot k_b > k] \cdot C_{\mathcal{M}}(b \cdot k_b, k, m),
\end{equation*}
where $[b \cdot k_b > k]$ is $1$ when $b \cdot k_b > k$ and $0$ otherwise. We assume here that costs are additive between \textbf{Stage 1} and \textbf{Stage 2}.

\begin{table}[htp]
    \centering
    \caption{Cost models and algorithms for exact \tk{}.}
    \label{tab:cost-models}
    \begin{tabular}{llll} \toprule
        Model                     & Algorithm                  & Cost                               & Residual                  \\\midrule
        Basic                     & -                          & $m \cdot n \cdot (\log_2 k + 1) $  & $+ \mathcal{O}(m \cdot k)$ \\
        \multirow{2}{*}{Serial}   & \method{PriorityQueue}     & $m \cdot n \cdot (3 k - 1) $       & $+ \mathcal{O}(m)$         \\
                                  & \method{RadixSelect}       & $m \cdot n \cdot (4\log_2 n + 4) $ & $+ \mathcal{O}(m \log n)$  \\
        \multirow{2}{*}{Parallel} & \method{ScanMax}           & $k \cdot (2\log_2 n + 3) $         & $+ \mathcal{O}(1)$         \\
                                  & \method{RadixSelect}       & $\log_2 n \cdot (2\log_2 n + 16) $ & $+ \mathcal{O}(1)$         \\\bottomrule
    \end{tabular}
\end{table}

\subsection{Basic cost model}

The basic cost model uses the asymptotic bound for top-$k$ based on a linear scan through the data, maintaining a min-heap \parencite{forsythe1964_heap} of the top-$k$ of the prefix that has been scanned. Note that there are asymptotically faster top-$k$ algorithms such as \method{QuickSelect} \parencite{hoare1961_quickselect}, however these are ill-suited to execution on highly parallel hardware, which is the focus of this work. Since this top-$k$ algorithm is $\mathcal{O}(n \log k)$ and we note that top-$1$ is readily achieved with an $\mathcal{O}(n)$ scan through the data, we set the cost $C_{\textsc{basic}} = m \cdot n \cdot (\log_2 k + 1)$.

\subsection{Serial and Parallel cost models}

In these cost models, we go beyond asymptotic evaluation, since the combination of multiple algorithms based on the minimum cost depends on the constant factors associated with each algorithm. To do this, we describe a simple abstract execution model, with the following assumptions:
\begin{itemize}
    \item Iterating through the input data linearly is free.
    \item Fixed-offset addressing is free.
    \item Bounds checking is free.
    \item If statements are free, but all branches are taken.
    \item Ignore small ``residual'' cost terms.
    \item Operations that cost $1$: \texttt{==}, \texttt{<}, \texttt{>}, \texttt{+}, \texttt{\&}, \texttt{|}, \texttt{<{}<},  \texttt{not}, \texttt{=} (arithmetic, logical, assignment).
    \item Operations that cost $2$: \texttt{+=}, \texttt{\&=}, \texttt{|=}, \texttt{\^{}=} (read-modify-write).
\end{itemize}
We then define two cost model variants:
\begin{itemize}
    \item \emph{Serial}, where the total cost is the sum of all operation costs.
    \item \emph{Parallel}, where the total cost is total cost of operations executed by a machine with infinite parallel threads.
\end{itemize}

We evaluate two algorithms for each cost model, one appropriate for small-$k$, another for large-$k$, with a rough correspondence to the priority queue and \method{RadixSelect} implementations that we have evaluated. Costs are outlined in \cref{tab:cost-models} and explained using illustrative Python code in \cref{code:serial-insertion,code:serial-radix,code:parallel-scan-max,code:parallel-radix}.

\begin{figure}
\begin{minted}{python}
def topk_insertion(data, k):
    topk = sorted(data[:k])                               # (ignore)
    for x in data:                                        # * n
        if x > topk[0]:                                   # | +1
            topk[0] = x                                   # | +1
        for j in range(1, k):                             # | * (k-1)
            if topk[j-1] > topk[j]:                       # | | +1
                topk[j-1], topk[j] = topk[j], topk[j-1]   # | | +2
    return topk
\end{minted}
    \caption{\method{PriorityQueue} (serial), with insertion sort, cost $m \cdot n \cdot (3 k - 1) + \mathcal{O}(m)$. The first sort can be merged into the loop.}
    \label{code:serial-insertion}
\end{figure}

\begin{figure}
\begin{minted}{python}
def topk_radix_select(data, k):
    # Find kth value
    kth_value, mask, count_gt = 0, 0, 0
    for r in range(31, -1, -1):          # * log(n)
        r_mask = 1 << r
        kth_value |= r_mask
        mask |= r_mask
        count_1 = 0
        for x, _ in data:                # | * n
            if x & mask == kth_value:    # | | +2
                count_1 += 1             # | | +2
        if count_gt + count_1 < k:
            kth_value ^= r_mask
            count_gt += count_1

    # Collect topk
    topk = [None] * k
    i = 0
    for x in data:             # * n
        if x[0] >= kth_value:  # | +1
            topk[i] = x        # | +1
            i += 1             # | +2
    return topk
\end{minted}
    \caption{\method{RadixSelect} (serial), cost $m \cdot n \cdot (4\log_2 n + 4) + \mathcal{O}(m \log n)$. Note that we assume a key length of $\log_2 n$, to uniquely identify $n$ elements; in practical scenarios, the key length is separate from $n$. If the $k$\textsuperscript{th} element may be tied, a second ``collect'' step may be necessary to ensure that only these tied elements are discarded.}
    \label{code:serial-radix}
\end{figure}

\begin{figure}
\begin{minted}{python}
def scan_argmax(data):
    a = list(range(len(data)))                   # +1
    for i in range(log2(len(data))):             # * log(n)
        a = [a[j]                                # | +1
             if j+2**i >= len(data) \
                or data[a[j+2**i]] < data[a[j]]  # | +1
             else a[j+2**i]
             for j in range(len(data))]
    return a[0]

def topk_scan_max(data, k):
    data = data.copy()
    topk = [None] * k
    for i in range(k):         # * k
        j = scan_argmax(data)  # | +2*log(n) + 1
        topk[i] = data[j]      # | +1
        data[j] = (-inf, 0)    # | +1
    return topk
\end{minted}
    \caption{\method{ScanMax} (parallel), cost $k \cdot (2\log_2 n + 3) + \mathcal{O}(1)$.}
    \label{code:parallel-scan-max}
\end{figure}

\begin{figure}
\begin{minted}{python}
def scan_cumsum(data):
    s = data.copy()                                    # +1
    for i in range(log2(len(s))):                      # * log(n)
        s = [s[j] + (s[j-2**i] if j-2**i >= 0 else 0)  # | +2
             for j in range(len(s))]
    return s

def topk_radix_select_parallel(data, k):
    # Find kth value
    kth_value, mask, count_gt = 0, 0, 0
    for r in range(31, -1, -1):                              # * log(n)
        r_mask = 1 << r                                      # | +1
        kth_value |= r_mask                                  # | +2
        mask |= r_mask                                       # | +2
        count_1s = [x & mask == kth_value for x, _ in data]  # | +2
        count_1 = scan_cumsum(count_1s)[-1]                  # | +2*log(n) + 1
        if count_gt + count_1 < k:                           # | +2
            kth_value ^= r_mask                              # | +2
            count_gt += count_1                              # | +2

    # Collect topk
    in_topk = [x >= kth_value for x, _ in data]
    offset = scan_cumsum(in_topk)                            # +2*log(n) (+ 1)
    topk = [None] * k
    for i in range(len(data)):                               # (in parallel)
        if in_topk[i]:
            topk[offset[i] - 1] = data[i]
    return topk
\end{minted}
    \caption{\method{RadixSelect} (parallel), cost $\log_2 n \cdot (2\log_2 n + 16) + \mathcal{O}(1)$.}
    \label{code:parallel-radix}
\end{figure}
\clearpage

\subsection{Analysing cost models}
\label{app:cost-models-analysis}
We compare the scaling trends predicted by our abstract cost models and the practical benchmarking results observed for \texttt{torch.topk}, which is based on \method{RadixSelect} and our CUDA implementation of a \method{PriorityQueue} using insertion sort.

\begin{figure}[p]
    \centering
    \includegraphics[width=\textwidth]{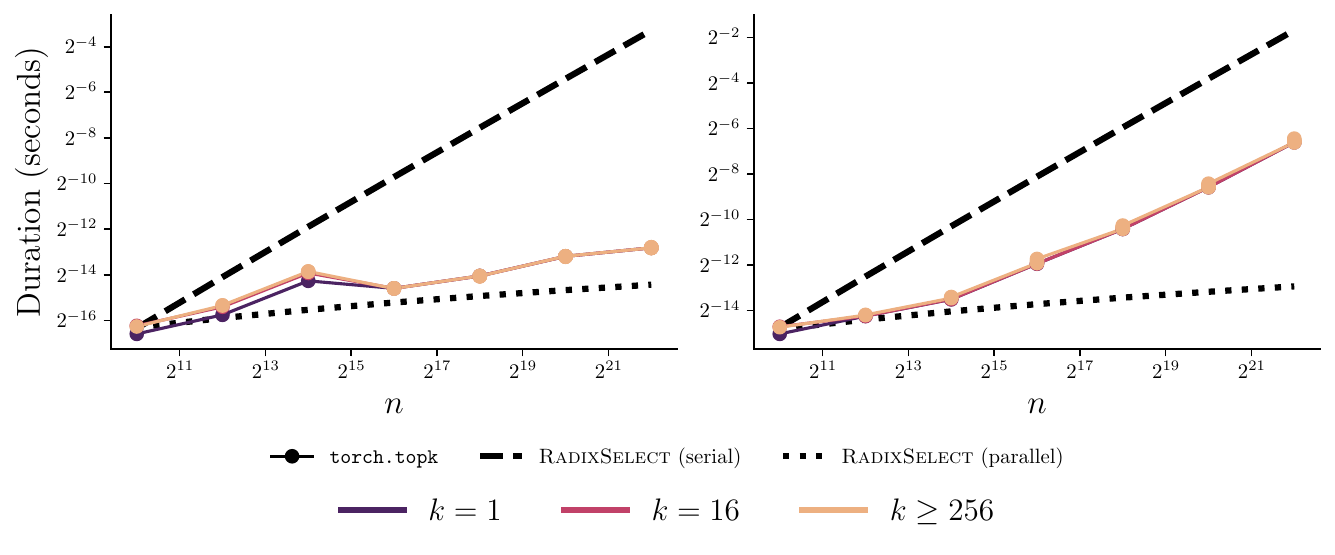}
    \caption{
    PyTorch \tk{} runtime as a function of $n$, $k$. \emph{Left:} Batch size $m=1$. \emph{Right:} $m=256$. The serial and parallel cost models have been aligned to match PyTorch at $n=2^{10}$. Practical runtime does not depend strongly on $k$, and follows the parallel cost model when batch size $m$ and input length $n$ are small, but then trends toward the serial cost model.
    }
    \label{fig:cost-model-torch}
\end{figure}

\Cref{fig:cost-model-torch} shows the performance of \texttt{torch.topk} as $n$ and $k$ vary, versus the corresponding cost model of \cref{tab:cost-models}. As expected, \method{RadixSelect} performance does not depend strongly on $k$, and behaves like the parallel cost model for small problems (small $m \cdot n$), then like the serial cost model for larger problems. This is to be expected of a GPU, which has resources to support a large finite number of parallel threads. When the available parallelism is under-utilised, implementations scale as if the number of threads is infinite. As the resources are exhausted, implementations scale based on the total amount of work.

\begin{figure}[p]
    \centering
    \includegraphics[width=\textwidth]{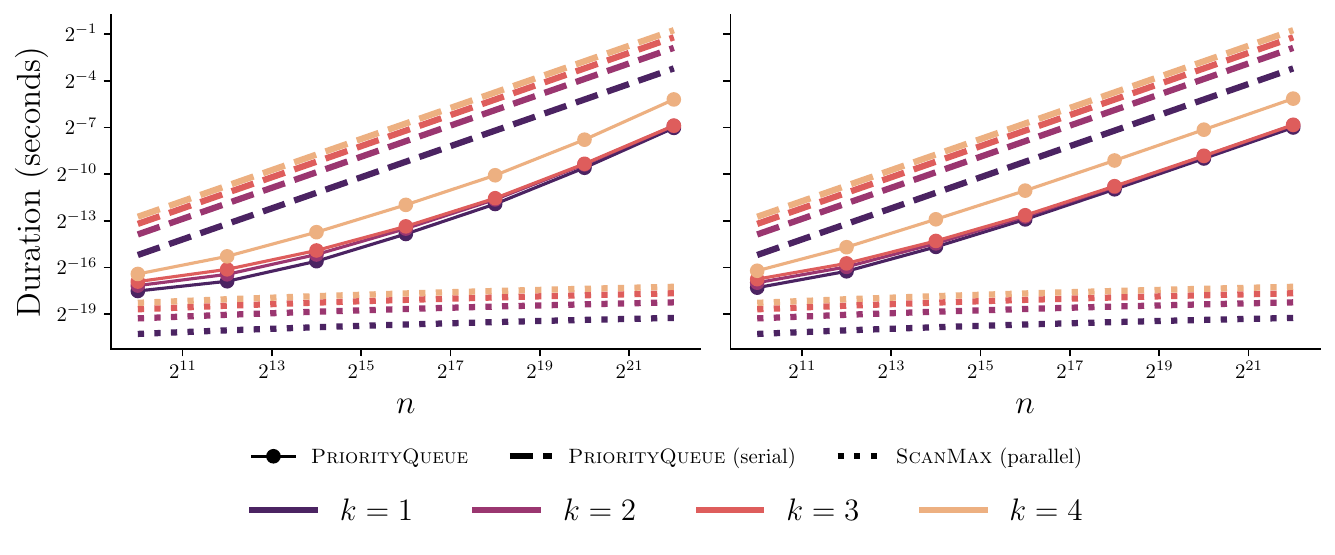}
    \caption{
    Our priority queue \tk{} runtime as a function of $n$, $k$. \emph{Left:} Batch size $m=1$. \emph{Right:} $m=256$. The serial and parallel cost models have been shifted vertically for sake of visual tracking, since they do not directly predict wall-clock time. Runtime scaling with $k$ is better than the predicted linear scaling (\cref{tab:cost-models}), especially for $k\in\{2,3\}$. At very small $n \cdot m$, runtime scaling with $n$ is close to the parallel model, but otherwise it scales as per the serial model.
    }
    \label{fig:cost-model-priority-queue}
\end{figure}

We perform a similar exercise for our implementation of \method{PriorityQueue} in \cref{fig:cost-model-priority-queue}. The serial cost model is generally a good predictor of runtime, but real runtime scales better than expected with $k$. This may be because the kernel is often bandwidth-bound w.r.t.\ reads from memory. In this case, there may be no additional cost to computing $k=2$ versus $k=1$, since the computation time is hidden by memory access.

\begin{figure}[tp]
    \centering
    \includegraphics[width=\textwidth]{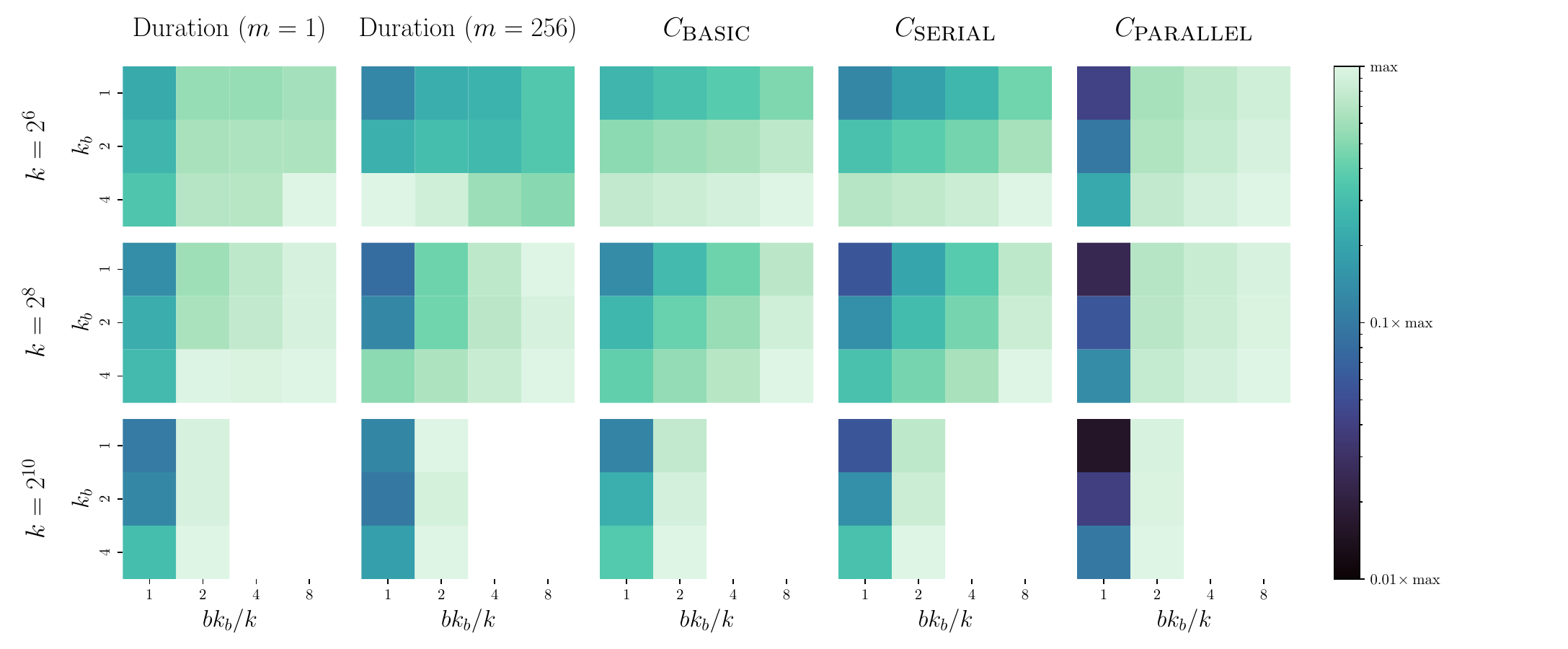}
    \includegraphics[width=\textwidth]{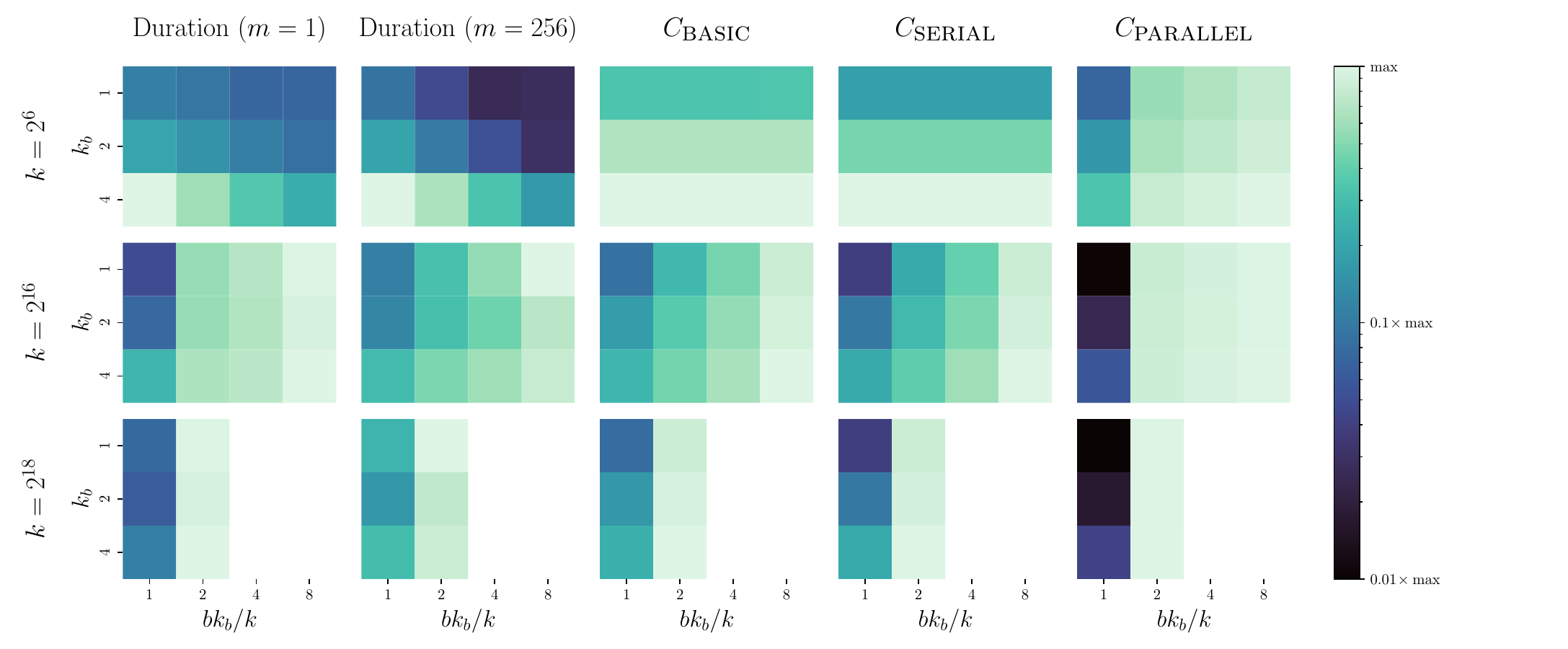}
    \caption{
    A comparison of the relative runtime of our approximate \tk{} implementation, against cost models $C$. \emph{Top:} $n=2^{12}$. \emph{Bottom:} $n=2^{20}$.
    These show that the basic or serial cost models are generally a reasonable match to practical runtime. One exception is that they generally underestimate the cost of having \textbf{Stage 2} (the step between $b \cdot k_b/k=1$ and $2$). The other notable exception is for $n=2^{20}, k=2^6$, where the runtime profile is inverted, with smaller $b\cdot k_b/k$ taking longer than larger $b\cdot k_b/k$.
    }
    \label{fig:cost-model-approx-topk-small-n}
\end{figure}

\section{Trade-off curves \label{appendix:tradeoff-curves}}

\begin{figure}
    \centering
    \includegraphics[width=\textwidth]{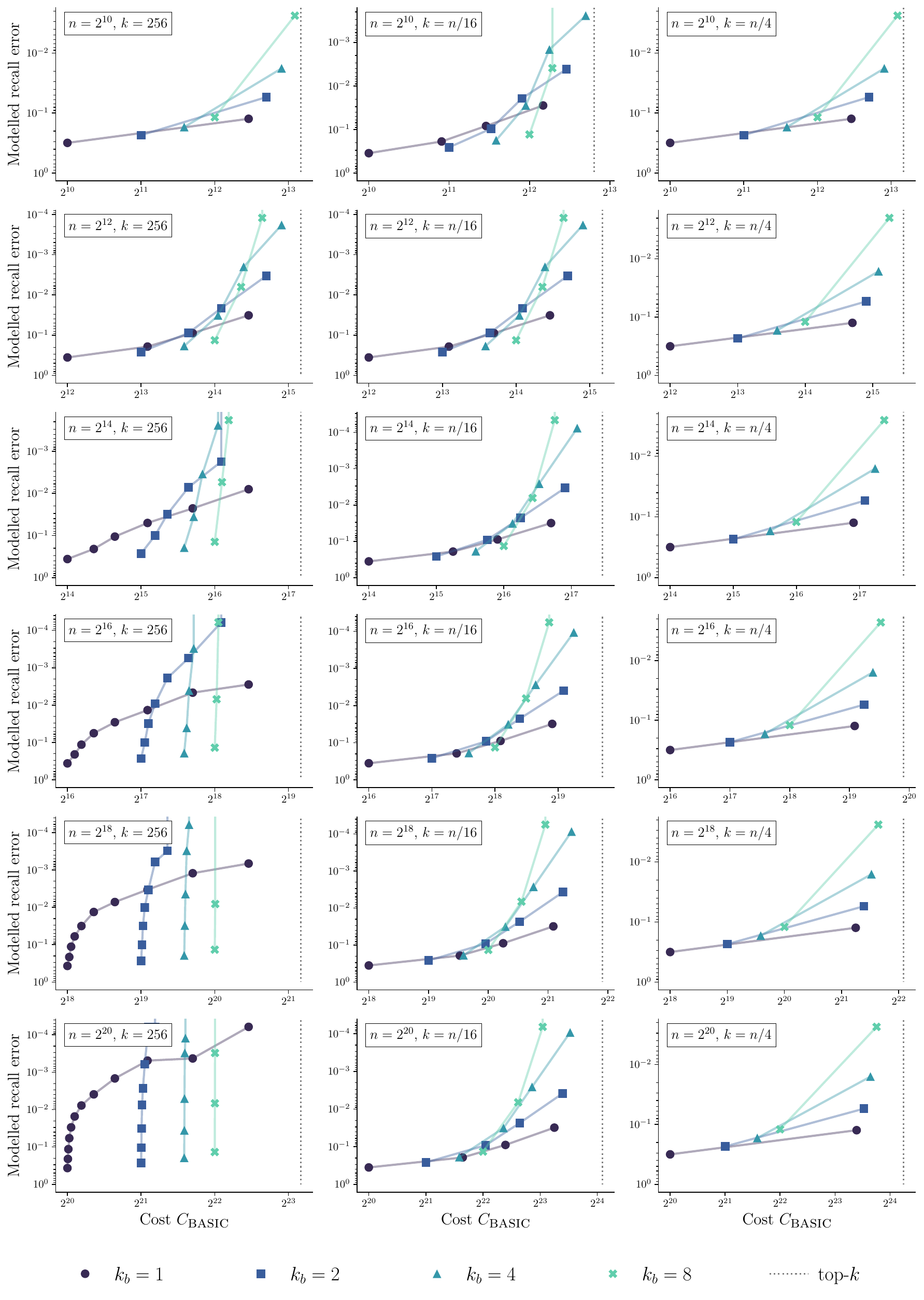}
    \caption{
    Theoretical trade-off curves under the basic cost model, as $n$, $k$, $k_b$ and $b\cdot k_b/k$ are varied. Columns show three different regimes based on $k$. Points along the curves correspond to different $b\cdot k_b/k$ ratios. \emph{Left:} small fixed $k=256$. \emph{Center:} moderate $k=n/16$. \emph{Right:} large $k=n/4$. Rows show increasing $n$.
    Figures in the bottom left demonstrate the small-$k$ regime, where $k_b=1$ should be preferred, and $b$ increased to reduce error. Figures in the center and right show that for large $k$, it is often better to increase $k_b>1$. In general, the dependence on $n$ is limited, given the ratio $k/n$.
    }
    \label{fig:theoretical-tradeoff-basic}
\end{figure}

\begin{figure}
    \centering
    \includegraphics[width=\textwidth]{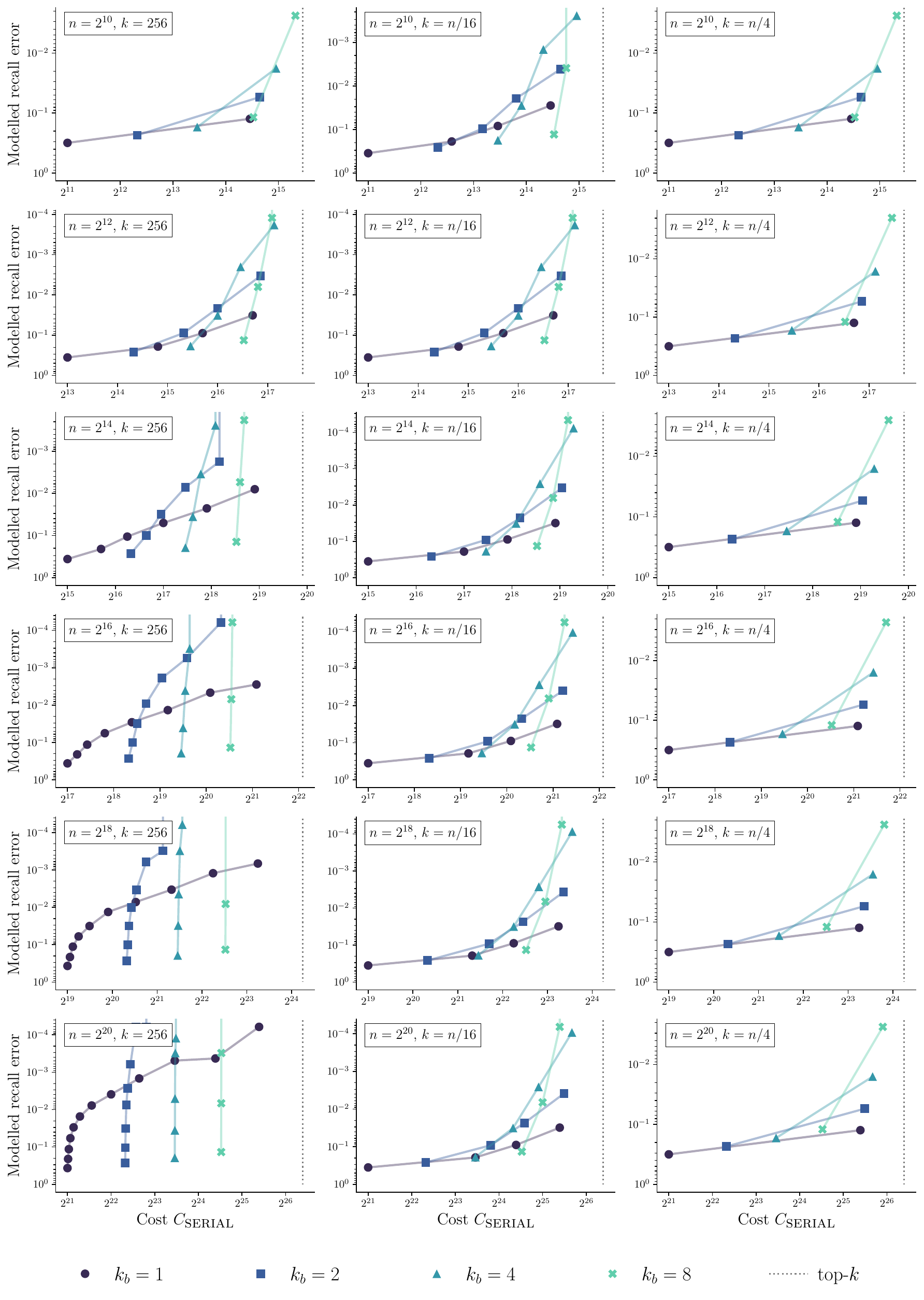}
    \caption{
    Theoretical trade-off curves under the serial cost model, as $n$, $k$, $k_b$ and $b\cdot k_b/k$ are varied. We observe that approximate \tk{} in this model scales similarly to that of the basic model; all observations made for \cref{fig:theoretical-tradeoff-basic} apply here too.
    }
    \label{fig:theoretical-tradeoff-serial}
\end{figure}

\begin{figure}
    \centering
    \includegraphics[width=\textwidth]{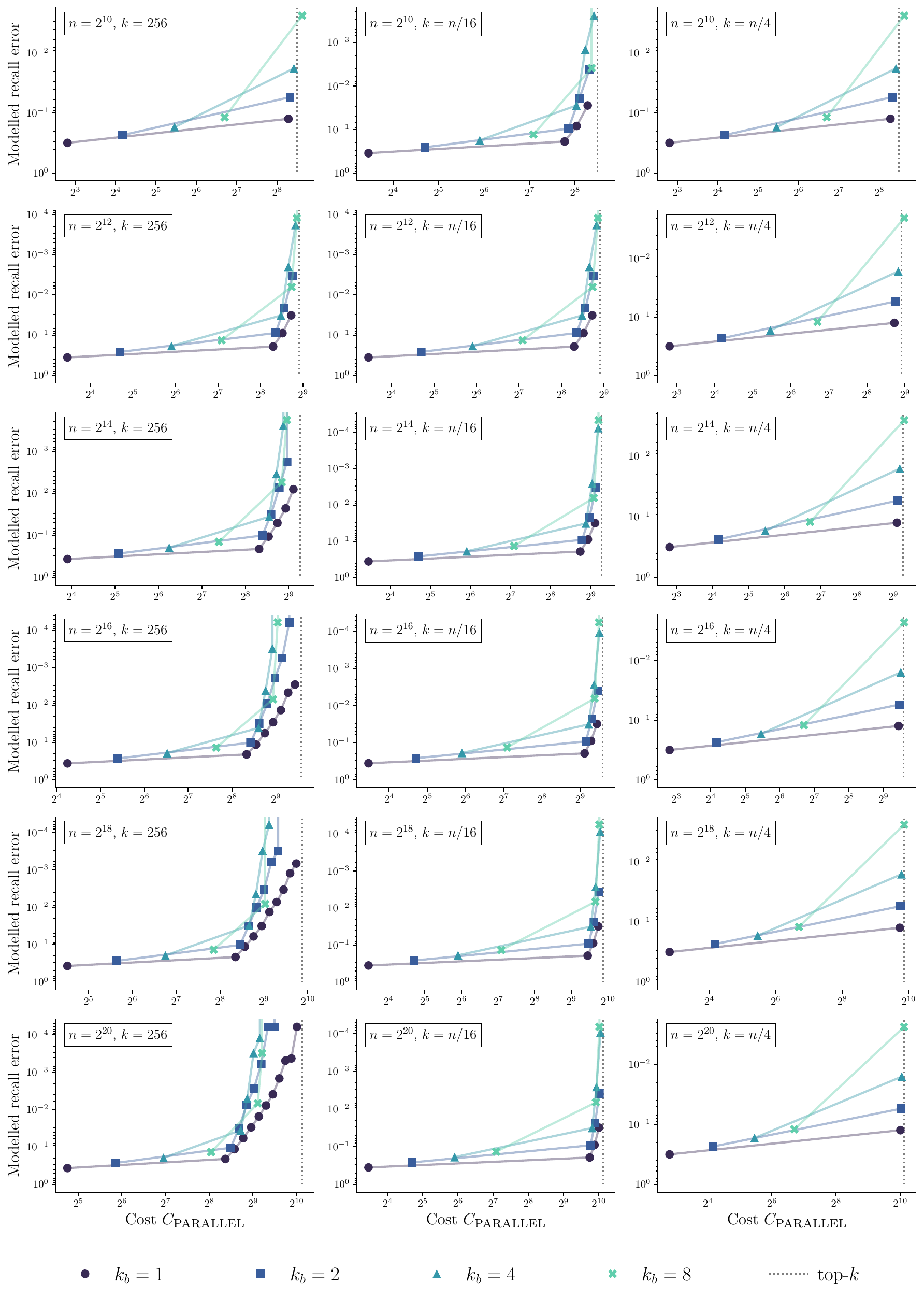}
    \caption{
    Theoretical trade-off curves under the parallel cost model, as $n$, $k$, $k_b$ and $b\cdot k_b/k$ are varied. The parallel cost model places a premium on the execution of \textbf{Stage 2}, which is avoided when $b\cdot k_b=k$ (for the leftmost points of each line). It is therefore optimal under this model to use $k_b$ to control error, even in the small-$k$ regime (left column). We note that it is hard to achieve substantial speed-ups with very low error under the parallel cost model.
    }
    \label{fig:theoretical-tradeoff-parallel}
\end{figure}

Combining our theoretical recall and cost models (\cref{app:recall-model,app:cost-models}), we have made a broad sweep of trade-off curves to help guide hyperparameter selection of $k_b$ and $b$. These are shown in \cref{fig:theoretical-tradeoff-basic,fig:theoretical-tradeoff-serial,fig:theoretical-tradeoff-parallel} for each cost model in turn. They motivate the following observations:

\begin{itemize}
    \item Low recall error and large speed-ups are available in the small-$k$ regime, roughly when $k/n \le 1/64$ (in both serial and basic cost models).
    \item In the small-$k$ regime, it is best to control error by increasing the number of buckets such that $b\cdot k_b>k$, keeping $k_b=1$. To achieve very low error, it eventually becomes optimal to increase $k_b>1$ too.
    \item In the large-$k$ regime, it is better to control error by increasing the per-bucket $k_b > 1$, while maintaining $b\cdot k_b=k$. This is because \textbf{Stage 2} can be very expensive, and it is often better to avoid it entirely.
    \item Substantial speed-ups are available in the parallel cost model, as long as \textbf{Stage 2} is not required. In this cost model, increasing $k_b>1$ should always be prioritised over increasing $b\cdot k_b>k$.
\end{itemize}


\end{document}